\documentclass{article}

\usepackage{natbib}
\usepackage{microtype}
\usepackage{graphicx}
\usepackage{subfigure}
\usepackage{booktabs} 
\usepackage{amsmath}
\usepackage{amssymb}
\usepackage{booktabs}
\usepackage{multirow}
\usepackage{multicol}
\usepackage{subfigure}
\usepackage[mathscr]{euscript}
\usepackage{dsfont}
\usepackage{color}
\usepackage{dutchcal}
\usepackage{amsfonts}
\usepackage{pgfplots}
\usepackage{colortbl}

\usepackage{hyperref}

\usepackage[accepted]{icml2024}

\usepackage{amsmath}
\usepackage{amssymb}
\usepackage{mathtools}
\usepackage{amsthm}

\usepackage[capitalize,noabbrev]{cleveref}

\theoremstyle{plain}

\theoremstyle{definition}

\theoremstyle{remark}

\usepackage[textsize=tiny]{todonotes}

\icmltitlerunning{Collapse-Aware Triplet Decoupling for Adversarially Robust Image Retrieval}

\pgfplotsset{compat=1.18}

\begin{document}

\twocolumn[
\icmltitle{Collapse-Aware Triplet Decoupling for Adversarially Robust Image Retrieval}

\begin{icmlauthorlist}
\icmlauthor{Qiwei Tian}{yyy}
\icmlauthor{Chenhao Lin}{yyy}
\icmlauthor{Zhengyu Zhao}{yyy}
\icmlauthor{Qian Li}{yyy}
\icmlauthor{Chao Shen}{yyy}

\end{icmlauthorlist}

\icmlaffiliation{yyy}{Xi'an JiaoTong University, Xi'an, China}

\icmlcorrespondingauthor{Chenhao Lin}{linchenhao@xjtu.edu.cn}
\icmlcorrespondingauthor{Chao Shen}{chaoshen@mail.xjtu.edu.cn}

\icmlkeywords{Image Retrieval, Model Collapse, Adversarial Defense, Robustness, Deep Metric Learning}

\vskip 0.3in
]


\printAffiliationsAndNotice{}  

\begin{abstract}
Adversarial training has achieved substantial performance in defending image retrieval against adversarial examples. However, existing studies in deep metric learning (DML) still suffer from two major limitations: \textit{weak adversary} and \textit{model collapse}. In this paper, we address these two limitations by proposing \textbf{C}ollapse-\textbf{A}ware \textbf{TRI}plet \textbf{DE}coupling (CA-TRIDE). Specifically, TRIDE yields a stronger adversary by spatially decoupling the perturbation targets into the anchor and the other candidates. Furthermore, CA prevents the consequential model collapse, based on a novel metric, collapseness, which is incorporated into the optimization of perturbation. We also identify two drawbacks of the existing robustness metric in image retrieval and propose a new metric for a more reasonable robustness evaluation. Extensive experiments on three datasets demonstrate that CA-TRIDE outperforms existing defense methods in both conventional and new metrics. \textit{Codes are available at} \hyperlink{https://github.com/michaeltian108/CA-TRIDE}{https://github.com/michaeltian108/CA-TRIDE}.
\end{abstract}

\section{Introduction}
\label{sec:intro}
\noindent 
Thanks to the massive image data and the development of Deep Neural Networks (DNNs), image retrieval has experienced substantial advancements.
Despite their effectiveness, deep image retrieval is known to be vulnerable to adversarial examples, i.e. test samples that cause erroneous model behaviour \citep{AA}.
Existing work on adversarial attacks against DNN-based image retrieval has explored deep metric learning (DML) models \citep{QAM, QAIR, DAIR, RA} and deep hashing models \citep{AA:hash1, AA:hash2}.

To defend DML models against adversarial examples, recent studies \citep{ACT,HM} have modified the commonly-used adversarial training \citep{PGD} from the domain of image classification into the domain of image retrieval.
However, these methods still suffer from two major limitations:

\begin{figure}[!t]

 \subfigure{
    \includegraphics[width=\columnwidth]{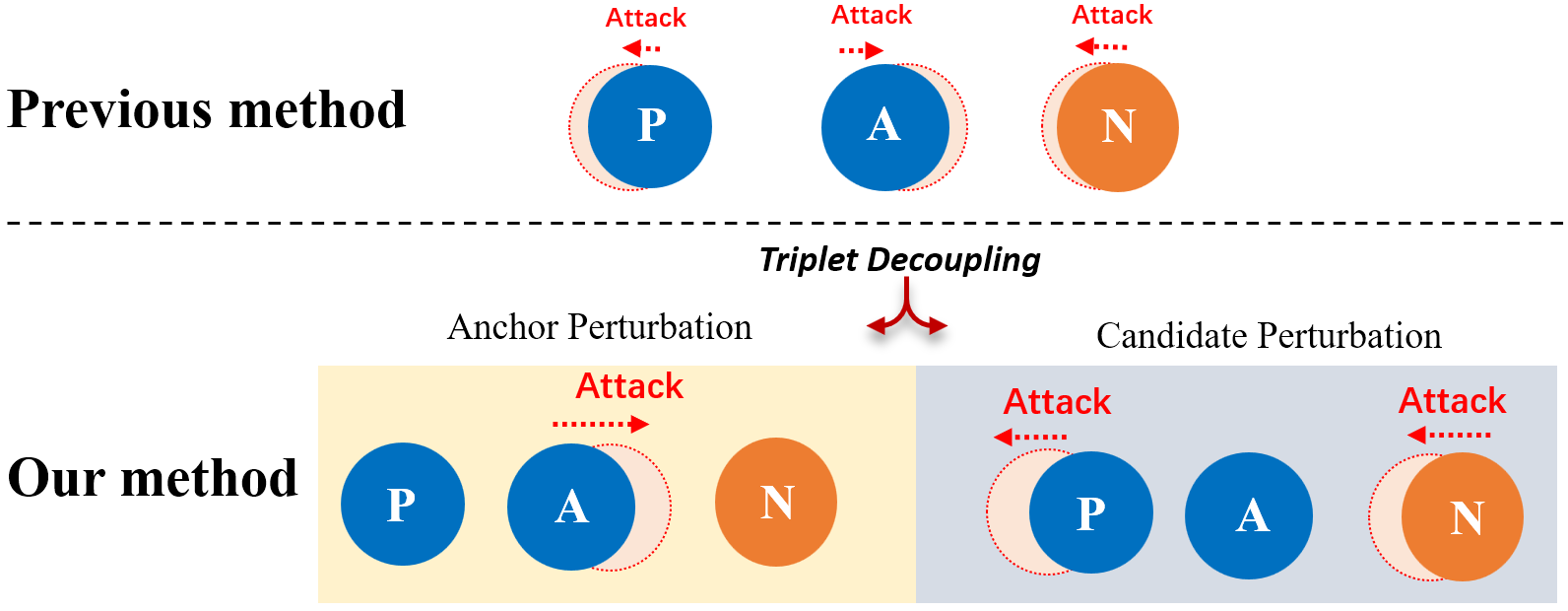}
    }
    
  \caption{\label{img_insufadv}
  Our defense vs. the previous defense. Specifically, we address the \textit{weak adversary} by decoupling the updates of perturbation into anchors (A) and positive (P)/negative (N) candidates to maximize embedding shifts.}

\end{figure}

\begin{itemize}
     \item \textbf {Weak adversary} is a widely overlooked issue since existing defenses \citep{HM, ACT} directly adopt the perturbation method from the image classification domain without fully exploiting the triplet structure, which consists of three components: positives, negatives, and anchors.
    As shown in Figure \ref{img_insufadv}, the existing method perturbs all three components simultaneously and allocates the desired embedding shifts among each component, as represented by the small red shadows, which leads to smaller average embedding shifts. 
    A detailed qualitative analysis is presented in Section \ref{IME} of the Appendix.

     \item \textbf{Model Collapse} is a notorious challenge in DML, impeding researchers from training models with hard samples \citep{MC}. A collapsed model embeds all samples disastrously close and cannot retrieve semantically similar examples appropriately. Model collapse becomes inevitable in adversarial training as it aims at increasing sample hardness.
     The current state-of-the-art adversarial training method, Hardness Manipulation \citep{HM}, avoids model collapse by restricting perturbation strength, consequently lowering its gained robustness. 
    
\end{itemize}

In this paper, we address the above limitations by proposing a novel adversarial training approach to robust image retrieval, called \textbf{C}ollapseness-\textbf{A}ware \textbf{TRI}plet \textbf{DE}coupling (\textbf{CA-TRIDE}). 
A comparison between our CA-TRIDE and the previous method is illustrated in Figure \ref{img_insufadv}. Generally, our CA-TRIDE can be applied to any triplet-based DML in a plug-and-play manner, without making major modifications.

First, our TRIDE aims to address the \textit{weak adversary} by decoupling the perturbation updates on the triplets (anchors, positive candidates, and negative candidates) into two phases: anchor perturbation (ANP) and candidate perturbation (CAP).
Specifically, in the ANP phase, only the anchor is perturbed, while in the CAP phase, the positive and negative candidates are perturbed.
By alternating between CAP and ANP, TRIDE incurs larger embedding shifts and yields a stronger adversary, as represented by the red shadows in Figure \ref{img_insufadv}.
Then, to tackle model collapse caused by our stronger adversary, CA is proposed to use a novel metric `collapseness' (denoted as $\mathcal{C}$) to capture intermediate model states, preventing impending model collapse by orienting the optimization of TRIDE through $\mathcal{C}$.

In sum, our paper makes the following three contributions:
\begin{itemize}
    \item We propose CA-TRIDE, a novel approach to adversarial robust image retrieval based on collapse-aware (CA) and triplet decoupling (TRIDE).
    In particular, a new metric called `collapseness' is proposed to capture the intermediate model states w.r.t. model collapse.

    \item We validate that CA-TRIDE addresses the two limitations in existing defense: \textit{weak adversary} and \textit{model collapse}, showing that TRIDE yields noticeably larger embedding shifts, more shrinkage in embedding space and lower separability, without causing model collapse.
    
    \item We identify two drawbacks of the commonly used robustness evaluation metric, ERS \citep{ACT}, and propose a new metric, ARS. 
    Experimental results on three datasets show that CA-TRIDE outperforms existing methods in both ERS and ARS by 4$\sim$5\%.
\end{itemize}

\begin{figure*}[!t]
    \centering
    \includegraphics[width=\textwidth]{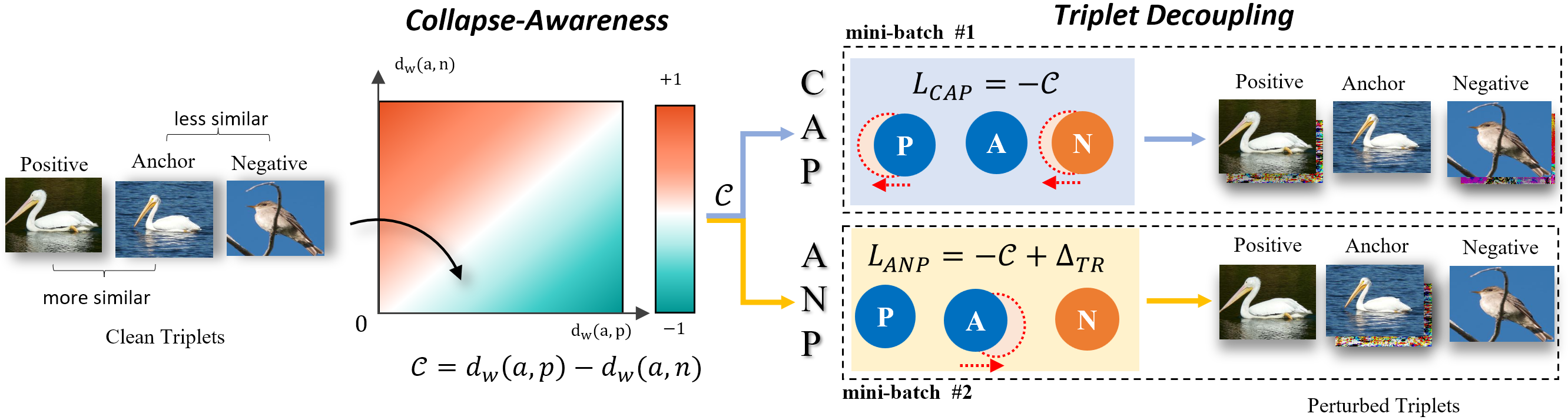}
        \vspace{-10pt}
        \caption{
    The working pipeline of our \textbf{CA-TRIDE} defense on a per mini-batch basis. (1) \textbf{Collapseness-Awareness (CA)} first calculates the proposed collapseness $\mathcal{C}$ on the clean triplets to capture model states and then incorporates $\mathcal{C}$ into subsequent perturbation optimization. (2) \textbf{Triplet Decoupling (TRIDE)} decouples the perturbation targets on the triplets into candidate perturbation (CAP) and anchor perturbation (ANP), which are alternatively implemented across mini-batches during adversarial training, starting with CAP.}

    \label{img_ov}
\end{figure*}

\vspace{-0.4cm}
\section{Related Work} \label{Related Works}

\subsection{Image Retrieval and Deep Metric Learning} Image retrieval \citep{IR} has been thriving as a surging amount of visual content has become available to the public. Deep metric learning (DML) is one of the popular methods to realize such tasks. 
Two main focuses of current work in DML are loss functions and data sampling methods, both of which have a crucial impact on image retrieval performance \cite{dmlcheck}. For loss functions, among other newly proposed losses, such as lifted structure loss \citep{lifted:ref}, N-pair loss \citep{Npair:ref} and Multi-Similarity loss \citep{Multsim:ref}, triplet loss \citep{trip:ref, trip:ref2} has been popular due to its simplicity and performance. For sampling methods, current research focuses on improving the diversity of samples within a batch to enhance the overall performance. Popular sampling strategies include random sampling, semi-hard sampling \citep{trip:ref}, soft-hard sampling \citep{revisit}, distance-weighted sampling \citep{dw}, etc. 
In this paper, we adopt the commonly used triplet loss for training DML models.

\subsection{Adversarial Attacks in Deep Metric Learning} 
Adversarial attacks (AAs) against image retrieval primarily seek to lower the R@1 of image retrieval \citep{QAM,TMA,LTM,PQAG}. These attacks, such as DAIR \citep{DAIR} and QAIR \citep{QAIR}, are mostly black-box and realized through repetitive queries to get a subset of the victim model's training set, which is subsequently used for optimizing AAs.
Universal adversarial perturbations (UAPs) have also been explored for image retrieval, which works by utilizing rank-wise relationships to increase the transferability of AAs \citep{UAP}. 
In addition, \citet{RA} explores a ranking attack that compromises the ranking results, i.e. improves or decreases the rank of certain candidates.

\subsection{Adversarial Defenses in Deep Metric Learning} Current DML defenses against AAs are relatively less explored.
Similar to common adversarial defense practices in image classification \citep{AT:ref1, AT:ref2, AT:ref3, AT:ref4, AT:ref5}, existing defense methods in DML \citep{ACT, HM} also rely on adversarial training with a triplet loss. Specifically, \citet{ACT} proposed an indirect perturbation method, anti-collapse triplet (ACT), which pulls positive and negative candidates in the triplet closer to make them harder to distinguish, which consequently causes limited robustness.
In their later work \citep{HM}, a direct adversary called hardness manipulation (HM) \citep{HM} was proposed and achieved a better result than ACT.
However, as a certain number of adversaries were found to cause the model to collapse, the authors had to resort to weak adversaries to prevent such an issue.
In this paper, we specifically address the model collapse and weak adversary problems in existing defenses with our proposed CA-TRIDE.

\section{Methodology} \label{Methodology}

\subsection{Preliminaries}

\noindent \textbf{Triplet Loss.} Given a triplet $\mathbb{T} = (\mathbf{A},\mathbf{P}, \mathbf{N})$, the triplet loss $L_{\mathcal{T}}$ is defined as follows:
 \begin{gather}
      \label{triplet}
     L_{\mathcal{T}}(\mathbb{T})= d(\mathbf{A},\mathbf{P}) - d(\mathbf{A},\mathbf{N}) + \beta_{\mathcal{T}} 
 \end{gather}
where $\mathbf{A}$, $\mathbf{P}$, and $\mathbf{N}$ represent the anchors, positive, and negative examples, respectively. $d(\cdot,\cdot)$ calculates the average distance between two samples of the given triplet. $\beta_{\mathcal{T}}$ is the margin that defines how far the model needs to separate $\mathbf{P}$ from $\mathbf{N}$, usually set within $[0.2,0.8]$.

$L_{\mathcal{T}}$ helps the model learn an embedding space that locates semantically similar examples as closely as possible. This is achieved by minimizing $\mathit{d(\mathbf{A},\mathbf{P})}$ and maximizing $\mathit{d(\mathbf{A},\mathbf{N})}$ until the margin $\beta_{\mathcal{T}}$ is met. Consequently, the goal of adversarial training is otherwise: adversarially maximizing $\mathit{d(\mathbf{A},\mathbf{P})}$ while minimizing $\mathit{d(\mathbf{A},\mathbf{N})}$. 

\noindent \textbf{Hardness.} Proposed in the recent work \citep{HM}, the hardness of a triplet $\mathbb{T}$ is defined as follows:
\begin{equation}
\label{eq_hardness}
H(\mathbf{A},\mathbf{P},\mathbf{N}) = d(\mathbf{A},\mathbf{P}) - d(\mathbf{A},\mathbf{N}),~~H \in [-2,2]
\end{equation} 
\noindent Hardness determines how difficult a triplet is for a DML model to distinguish. A negative $H$ signifies easy triplets, while a positive $H$ denotes hard triplets. The magnitude of $H$ determines how hard or easy the triplet is. 

\noindent \textbf{Adversarial Training (AT) in DML.} In contrast to the standard training, the goal of AT is to learn a robust model on the adversarial triplet $\tilde{\mathbb{T}}$, which is acquired by adding perturbations ${\delta}$ into the clean triplet $\tilde{\mathbb{T}}=(\Tilde{\mathbf{A}},\Tilde{\mathbf{P}},\Tilde{\mathbf{N}}) = (\mathbf{A} +{\delta}, \mathbf{P}+{\delta}, \mathbf{N}+{\delta}$).

Specifically, the perturbation ${\delta}$ is optimized by maximizing hardness $H$:
\begin{equation}
\label{l_sip}
\arg \max_{{\delta}} H(\Tilde{\mathbf{A}},\Tilde{\mathbf{P}},\Tilde{\mathbf{N}})
\end{equation}

where $\delta$ is bounded by an $l_p$ norm $\epsilon$ for imperceptibility. The specific methods for maximizing $H$ could vary across different methods. For example, the afore-mentioned HM uses $L_{HM} = ||H_D - \tilde{H}_O||_2^2$ to generate perturbation to increase orginal hardness from $H_O$ to destination hardness $H_{D}$. 

After the generation of perturbed triplets, these triplets are used for training an adversarially robust model ${\Theta}$ by optimizing:
\begin{equation}
    \label{l_sip_eta}
    \arg\min_{{\theta}}L_{\mathcal{T}}(\Tilde{\mathbf{A}},\Tilde{\mathbf{P}},\Tilde{\mathbf{N}};\Theta)
\end{equation}

Our CA-TRIDE is built on AT, as shown in the per mini-batch working pipeline in Figure \ref{img_ov}. Overall, given a mini-batch of triplets, CA first evaluates model states on clean triplets using collapseness $\mathcal{C}$. The calculated collapseness is then fed into subsequent TRIDE to incorporate collapse awareness for the optimization of perturbations to prevent model collapse.
Specifically, TRIDE starts with CAP and alternates between CAP and ANP to perturb the triplets with the dedicated adversarial losses, $L_{CAP}$ and $L_{ANP}$. These perturbed triplets are then fed into the model for training, followed by another mini-batch of CA-TRIDE until the training is finished. The full process of generating perturbed triplets using CA-TRIDE is described in Algorithm \ref{alg}. {Note that our CA-TRIDE does not incur any extra epochs but only changes the number of perturbed samples per epoch.}
 
\subsection{Collapse-Aware Adversary (CA)} \label{CLA}
To incorporate collapse awareness in the optimization of perturbation and prevent model collapse, we first introduce \textit{collapseness} as a novel metric to capture model states proactively and accurately.

\noindent \textbf{Collapseness.} Hard triplets (i.e. $H>0$) are considered the main cause of model collapse due to their difficulty in optimization, according to current research \citep{MC, trip:ref}. Furthermore, the collapsing speed and severity depend on how many and how hard these triplets are. Although an intuitive practice is to use $H$ as a metric, it is not a feasible option as H only `reports' model collapse after it occurs and cannot `forecast' it beforehand, not to mention capturing an impending collapse. Thus, to guide subsequent counter-measurement against model collapse, we propose \textit{Collapsenss} as a proactive metric for model states evaluation, denoted as $\mathcal{C}$:
\begin{equation}
    \label{eq_c}
     \mathcal{C} (\mathbf{A},\mathbf{P},\mathbf{N})= d_{\omega}(\mathbf{A}, \mathbf{P}) - d_{\omega}(\mathbf{A}, \mathbf{N})
\end{equation}
$d_{\omega}(\cdot,\cdot)$ is a weighting function focusing on anchor-proximity samples. which is calculated as follows:
\begin{equation}
    \label{dw}
     d_{\omega}(\mathbf{A}, \mathbf{P})= \frac{ \sum_{i}^{\mathbf{A,P}}{\big (w_{p_{i}}\cdot d(a_{i}, p_{i})\big)}}
     {\sum_{i}^{\mathbf{P}}w_{p_{i}}} 
\end{equation}
\begin{equation}
\label{wpi}
w_{p_{i}} = exp\Big(-\lambda\big( d(a_{i},p_{i}) -
\min_{\forall a_{i} \in \textbf{A},p_{i} \in \textbf{P}}d(a_{i},p_{i})\big)\Big)
\end{equation}
\noindent where $\lambda$ is the attention factor to determine how much attention $\mathcal{C}$ should pay to anchor-proximity samples, and $exp(\cdot)$ is conventionally adopted to map all inputs into $[0,1]$, with a larger weight for closer samples and a smaller value for others. $d_{\omega}(\mathbf{A}, \mathbf{N})$ can be calculated similarly. \textbf{Positive $\mathcal{C}$ signals an impending model collapse, while negative $\mathcal{C}$ suggests a moderate hardness for the model.}

\begin{algorithm}[!t]
\caption{Generating Adversarial Triplets in \textbf{CA-TRIDE}}
\label{alg}
\begin{algorithmic}[1]
\REQUIRE Clean triplets $\mathbb{T} = (\mathbf{A},\mathbf{P},\mathbf{N})$, maximum PGD steps $M$, PGD step size $\alpha$
\ENSURE Adversarial Triplet $\tilde{\mathbb{T}}$
\STATE Initialize $\tilde{\mathbb{T}}_0 \gets \mathbb{T}$
\IF{CAP}
    \FOR{$i \gets 1$ to $M$}
        \STATE $\delta_i \gets \alpha \nabla_{\delta} L_{CAP}(\tilde{\mathbb{T}}_{i-1} )$
        \STATE $\tilde{\mathbb{T}}_i \gets \big(\mathbf{A},\tilde{\mathbf{P}}_{i-1} + \delta_{i}, \tilde{\mathbf{N}}_{i-1} + \delta_{i}\big)$
    \ENDFOR
\ELSIF{ANP}
    \FOR{$i \gets 1$ to $M$}
        \STATE $\delta_{i} \gets \alpha \nabla_{\delta} L_{ANP}(\tilde{\mathbb{T}}_{i-1} )$
        \STATE $\tilde{\mathbb{T}}_i \gets (\tilde{\mathbf{A}}_{i-1} + \delta_{{i}},\mathbf{P}, \mathbf{N})$
    \ENDFOR
\ENDIF
\STATE \textbf{return} $\tilde{\mathbb{T}} \gets \tilde{\mathbb{T}}_M$
\end{algorithmic}  
\end{algorithm}

As a cause of model collapse \citep{MC}, our $\mathcal{C}$ can detect undergoing model collapse and track the model state proactively by focusing more on neighbouring samples. We also evaluate both $H$ and $\mathcal{C}$ w.r.t. tracking model states to demonstrate the superiority of $\mathcal{C}$, presented in Section \ref{ablation}.

\noindent \textbf{Collapse-aware adversary.}
With our proposed $\mathcal{C}$, the adversarial goal of our subsequent perturbation optimization shifts from directing maximizing $H$ into maximizing $\mathcal{C}$:
\begin{flalign}
\label{l_colo}
    \arg \max_{\delta} \mathcal{C} (\Tilde{\mathbf{A}},\Tilde{\mathbf{P}},\Tilde{\mathbf{N}})
\end{flalign}

The definition of our collapse-aware adversary is intuitive as it utilizes the proposed $\mathcal{C}$ to optimize its adversary, while $\mathcal{C}$ also provides feedback for the optimization of the adversary to keep it aware of how well the model handles these perturbed samples.

In general, the maximization goal in Equation \ref{l_colo} determines the strength of AT. Since our primary goal of introducing a collapse-aware adversary is to prevent model collapse during training, we intuitively keep $\mathcal{C} \leq 0$ during the optimization of all perturbations to ensure that the adversary does not cause a severe model collapse while remaining reasonably strong to bring sufficient adversarial robustness.

\subsection{Triplet Decoupling (TRIDE)} \label{PSe} 
\noindent To address the weak adversary, we propose a new AT paradigm dubbed TRIDE to decouple the current perturbation method into two stages, candidate perturbation (CAP) and anchor perturbation (ANP), as shown in Figure \ref{img_insufadv}. Combining with our collapse-aware adversary, Equation \ref{l_colo} can be rewritten as:
\begin{flalign}
\label{l_colotride_r}
     \arg\max_{\delta} \begin{cases} 
     \mathcal{C}(\Tilde{\mathbf{A}}, \mathbf{P}, \mathbf{N}), &ANP  \\
      \mathcal{C}(\mathbf{A},\Tilde{\mathbf{P}}, \Tilde{\mathbf{N}}), &CAP
    \end{cases}
\end{flalign}
As depicted in Figure \ref{img_ov}, CAP and ANP are alternated mini-batch-wise, i.e. one perturbation method per mini-batch. 

\begin{figure}[!t]
    \centering
    \begin{minipage}{\linewidth} 
        \centering

        \subfigure[1][${\delta}= \frac{\Delta}{2}$]{\label{CAP_demo}
            \includegraphics[width=0.3\linewidth]{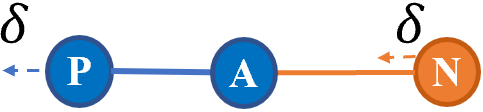}}
        \subfigure[2][${\delta}= \frac{\Delta}{2}$]{\label{ANP_demo}
            \includegraphics[width=0.3\linewidth]{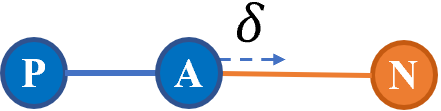}}
        \subfigure[3][${\delta} = \frac{\Delta}{4}$]{\label{SIP_demo}
            \includegraphics[width=0.3\linewidth]{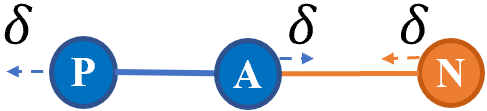}}
        \end{minipage}
    \caption{\label{TRIDE_demon}{An illustration of how the same desired hardness change $\Delta$ is transferred into sample-wise perturbation $\delta$ using different perturbation methods}: (a) CAP, (b) ANP, and (c) existing methods (i.e. HM and ACT). Our CAP and ANP double the averaged embedding shift in this specific case.    }
\end{figure}

Given a desired hardness change $\Delta$, the generated perturbation $\delta$ varies according to perturbation methods. Besides, the angular relationships between samples could also influence $\delta$. To eliminate the influence of angle and provide intuition on how perturbation methods impact $\delta$, we use a special case in Figure \ref{TRIDE_demon} to illustrate the difference between CAP, ANP, and the existing method. Analysis of more general cases can be found in Section \ref{IME} of the Appendix.

We now introduce the adversarial loss for CAP and ANP, which is designed following the inherent traits of each stage.

\noindent \textbf{Candidate perturbation (CAP).}
CAP perturbs all candidates (positives and negatives) while keeping the anchors fixed during perturbation optimization. $L_{CAP}$ is intuitively designed to maximize the aforementioned `collapsness' $\mathcal{C}$ in Equation \ref{eq_c}, by minimizing the following loss function:

\begin{flalign}
\label{lcap} 
    L_{CAP} = \max \big(-\mathcal{C},0\big) 
\end{flalign}
where the $\max(\cdot)$ term clips the loss to 0 to terminate the optimization once  $\mathcal{C}\geq0$. In this way, the optimization of CAP becomes collapse-aware and properly stops once the adversary becomes unacceptably strong for the model, thus preventing models from confronting severe model collapse.

CAP corrupts the global embedding (all negatives and positives) and helps the model defend against Ranking attacks (i.e. manipulating the ranks of one or more samples) by enhancing \textbf{global robustness}.

\noindent \textbf{Anchor perturbation (ANP).} ANP directly pushes the anchor away from the positives and towards the negatives. Similarly to CAP, the goal of $L_{ANP}$ is also to maximize $\mathcal{C}$, given below:
\begin{flalign}
\label{lanp} 
    L_{ANP} = \max \big(-\mathcal{C} + \Delta_{TR},0\big) 
\end{flalign}
ANP specializes in efficiently corrupting the local (anchor-proximity) embeddings, making it a practical implementation for black-box attacks against retrieval performance \citep{DAIR, QAIR}, i.e. Recall attacks. We propose a top-rank pair, consisting of a top-rank term $\Delta_{TR}$ and a top-rank triplet $L_{TR}$, to fully exploit such traits and help models acquire better resistance against such attacks through stronger \textbf{local} (top-rank) \textbf{robustness}.

\noindent  \textbf{Top-rank pair.}
The top-rank term $\Delta_{TR}$ is designed to push anchors towards the closest negatives further and maximize the embedding shift of anchors as $\mathcal{C}$ approaches 0, to reinforce local corruption:
\begin{equation}
    \label{ldtr} 
    \Delta_{TR} =e^{\max(\mathcal{C},0)} \big(d(\mathbf{A},\mathbf{N_{\upsilon}}) - d( \mathbf{A},\mathbf{A_0})\big) 
\end{equation}
where $\mathbf{N_{\upsilon}}$ represents the top half of negatives, ranked by their distances to anchors, and $\mathbf{A_0}$ stands for original unperturbed anchors. The $exp(\cdot)$ term ensures $\Delta_{TR}$ only kicks in as $\mathcal{C}$ approaches 0, i.e. $\mathbf{A}$ approaches $\mathbf{N}$. 

The top-rank triplet $L_{TR}$ is similarly defined using the triplet loss $L_{\mathcal{T}}$, pairing with $\Delta_{TR}$ to help models capture locality:
\begin{equation}
\label{drank}
L_{TR} =  \gamma \big(d(\mathbf{A}, \mathbf{P_{\upsilon}})- d(\mathbf{A},\mathbf{N_{\upsilon}}) + \beta_{TR} \big) 
\end{equation}
where $\gamma$ is a pre-determined coefficient, and $\mathbf{P_{\upsilon}} $ is similarly defined as $\mathbf{N_{\upsilon}}$, \textit{i.e.}, the top half of positives, ranked by their distances to the anchor.

In sum, our top-rank pair, $\Delta_{TR}$ and $L_{TR}$, works collaboratively to further boost the local robustness of models against prevailing black-box attacks deployed on anchors (queries) \citep{DAIR, QAIR}. $\Delta_{TR}$ reinforces local corruption for stronger, more targeted adversaries, while $L_{TR}$ helps the model to capture locality.

\noindent \textbf{Model training.}
As delineated in Equation \ref{l_sip_eta}, perturbed triplets generated through TRIDE are subsequently used for training a robust model by optimizing:
\begin{flalign}
\label{l_colotride}
     \arg\min_{\Theta} \begin{cases} 
     L_{\mathcal{T}}(\Tilde{\mathbf{A}}, \mathbf{P}, \mathbf{N};\Theta)+ L_{TR}, &ANP  \\
      L_{\mathcal{T}}(\mathbf{A},\Tilde{\mathbf{P}}, \Tilde{\mathbf{N}};\Theta), &CAP
    \end{cases}
\end{flalign}
In general, our TRIDE follows the lead of $\mathcal{C}$ to push the model to its limit by alternatively perturbing the triplets through CAP and ANP, enhancing its \textbf{global robustness} and \textbf{local robustness} respectively, without incurring disastrous model collapse.
More implementation details can be found in Appendix \ref{imp}.

\subsection{Robustness evaluation metrics}\label{AttackMetrics} 
\textbf{{Existing metrics.}} Existing works use ERS as the robustness evaluation metric, including 10 attacks for evaluation (details can be found in \citep{ACT}), which we categorize into \textbf{Ranking attacks} and \textbf{Recall attacks}. Ranking attacks, such as CA+ and CA-, evaluate the global robustness (i.e. overall ranking), while Recall attacks focus on evaluating local robustness by corrupting the proximity of anchors to lower overall retrieval performance (i.e. R@k).
\begin{table*}[!thpb]
 \caption{\label{tabl_AR}Robustness of ACT, HM, and our CA-TRIDE under our proposed metric, Adversarial Resistance Score (ARS).}
    \footnotesize
    \small
    \centering
   \resizebox{\linewidth}{!}{
   \begin{tabular}{c|c|c|c|c|c|c|cccc|cccc|c}
    \toprule
   \multirow{2}*{{Dataset}}  &  {Defense} & PGD & \multicolumn{4}{c|}{Benign Example Evaluation } & \multicolumn{8}{c|}{Adversarial Example Evaluation (ARS scores)}  &  {Overall} \\
    \cmidrule{4-7}   \cmidrule{8-15}
   & {Method} & steps & {R@1} $\uparrow$ & {R@2} $\uparrow$ & {mAP} $\uparrow$ & {NMI} $\uparrow$  & CA+ $\uparrow$ & CA- $\uparrow$ & QA+ $\uparrow$ & QA- $\uparrow$ & ES:R $\uparrow$ & LTM $\uparrow$ & GTM $\uparrow$ & GTT $\uparrow$ & ARS (\%) $\uparrow$\\ 
   
    \cmidrule(lr){1-16}
     \rowcolor{gray!20} \cellcolor{white}\multirow{4}*{CUB} 
    & N/A   & N/A            & {58.9}  & {66.4}  &  {26.1}  &  {59.5} &   3.3           & 0.0                &  0.0               &   0.0              &    0.0             &     0.0           &    23.9          & {0.0}          &        3.5     \\
    & ACT   & 32             & 27.5    & 38.2    &  12.2    &   43.0  &  31.0           &  62.9              &  30.2              &   68.5             &   40.3             &    34.2           &    54.2          &  1.0           &        40.3    \\
    & HM    & 32             & {34.9}  & 45.0    &  19.8    &   47.1  &  31.0           &  62.9              &  33.2              &   69.8             &   51.3             &    47.9           &   \textbf{78.2}  &  2.9           &        47.2    \\
    & Ours  &\textbf{ 16 }   & {34.9}  &  45.1   &  19.6    & 45.6    &\textbf{32.6}    &  \textbf{68.5}     &   \textbf{41.8}    &   \textbf{79.2}    & \textbf{61.9}      &  \textbf{59.0}    &   64.8           &   \textbf{5.1} &  \textbf{51.6} \\  
    \cmidrule(lr){1-16}
  \rowcolor{gray!20} \cellcolor{white} \multirow{4}*{CARS} 
    & N/A   &    N/A         &  {63.2} & { 75.3} & {36.6}   &  {55.6} &       0.4       &      0.0           & 0.0                & 3.6                &0.0                 & 0.0               & 21.2             & {0.0}          &       {2.8}     \\
    & ACT   &     32         &   43.4  &  54.6   &   11.8   &  42.9   &        36.0     &       68.4         &   35.0               &   70.2             &   37.6             &    35.3           &   47.7           &  1.6           &       41.4      \\
    & HM    &     32         &   60.2  &  71.6   &   33.9   &  51.2   & \textbf{38.6}   &      74.8          &   39.2             &    75.1            &   50.3             &    61.0           &  \textbf{ 76.4 } &        8.8     &       52.9      \\
    & Ours  &   \textbf{16 } &  {60.7} &  71.2   &  34.6    &  49.4   &         36.0    &   \textbf{81.0}    &   \textbf{47.0}    & \textbf{87.5}      &\textbf{64.4}       &   \textbf{66.9}   &   60.8           &  \textbf{13.7} &   \textbf{57.2} \\
  \cmidrule(lr){1-16}
  \rowcolor{gray!20} \cellcolor{white} \multirow{4}*{SOP}  
    & N/A   &      N/A       & {62.9}  & {68.5} &  {39.2}   & {87.4}  &         0.2     &        0.6         &       0.3          &      0.9           &        0.0         &       0.0         &       10.0       &        0.0     &       {1.5}     \\
    & ACT   &      32        &   47.5  &  52.6  &    25.5   & 84.9    &        48.2     &        90.4        &       45.4         &      91.5          &        44.6        &       45.5        &       58.5       &       15.3     &       54.9      \\
    & HM    &      32        &   46.8  &  51.7  &    24.5   & 84.7    &        64.0     &        96.8        &       67.4         &   \textbf{98.0}    &        83.5        &       85.0        &       81.0       &        45.6    &       77.7      \\
    & Ours  &   \textbf{16}  &  {48.3} &  53.3  &   25.9    & 84.9    &\textbf{65.8}    & \textbf{ 97.1}     &  \textbf{71.4}     &      {97.9}        & \textbf{89.4 }     &    \textbf{93.4}  & \textbf{82.4}    &  \textbf{53.1} &   \textbf{81.3} \\
    \bottomrule
    \end{tabular}
    }
    
\end{table*}

\begin{table*}[!t]
    \caption{\label{tabl_ERS}Robustness of ACT, HM, and our CA-TRIDE under the conventional metric, Empirical Robustness Score (ERS) \textcolor{black}{\cite{ACT}}.}
    \footnotesize
    \small
    \centering
   \resizebox{\linewidth}{!}{
   \begin{tabular}{c|c|c|c|c|c|c|cccccccccc|c}
    \toprule
   \multirow{2}*{{Dataset}} & {Defense} & PGD &  \multicolumn{4}{c|}{Benign Example Evaluation} & \multicolumn{10}{c|}{Adversarial Example Evaluation (ERS scores)} &  {Overall} \\
   \cmidrule{4-7}
   \cmidrule{8-17}
   &   {Method}  &  steps & {R@1} $\uparrow$ & {R@2} $\uparrow$ & {mAP} $\uparrow$ & {NMI} $\uparrow$ & CA+ $\uparrow$ & CA- $\downarrow$ & QA+ $\uparrow$ & QA- $\downarrow$ & {TMA} $\downarrow$ &{ES:D} $\downarrow$ & ES:R $\uparrow$ & LTM $\uparrow$ & GTM $\uparrow$ & GTT $\uparrow$ & {{ERS}$\uparrow$}\\ 
     
    \cmidrule(lr){1-18}
    
    \rowcolor{gray!20} \cellcolor{white}\multirow{4}*{CUB} 
        & N/A   &      N/A      & {58.9}  & {66.4}  &  {26.1}  & {59.5} &     0.0         &         100        &        0.0         &       99.9         &      0.883        &   1.76         &           0.0      &          0.0       &    14.1     &  {0.0}     &       {3.8}     \\
        & ACT   &      32       & 27.5    & 38.2    &  12.2    & 43.0   &     15.5        &         37.7       &       15.1         &       32.2         &    \textbf{0.47}  &   0.82         &           11.1     &          9.4       &    14.9     &   1.0      &        33.9     \\
        & HM    &      32       & 34.9    & 45.0    &  19.8    & 47.1   &     15.5        &         37.7       &       16.6         &       30.9         &       0.75        &   0.50         &           17.9     &          16.7      &\textbf{27.3}&   2.9      &        36.0     \\
        & Ours  &  \textbf{16}  & 34.9    &  45.1   &  19.6    & 45.6   & \textbf{16.7}   &    \textbf{31.1}   &   \textbf{20.9}    &  \textbf{ 21.1 }   &      0.97         &\textbf{0.16}   &    \textbf{21.6}   &    \textbf{20.6}   &    22.6     &\textbf{5.1}& \textbf{38.6}   \\
    \cmidrule(lr){1-18}
   \rowcolor{gray!20} \cellcolor{white} \multirow{4}*{CARS} 
        & N/A   &       N/A     & {63.2}  & { 75.3} & {36.6}  & {55.6}  &       0.2       &        100.0       &         0.1        &       97.3         &       0.87       &     1.82        &         0.0        &         0.0         &    13.4     &       {0.0}      &       {3.6}    \\
        & ACT   &       32      &  43.4   &  54.6   &   11.8  & 42.9    &       18.0      &         32.3       &        17.5        &       30.5         &  \textbf{0.38}   &     0.76        &         16.3       &         15.3        &    20.7     &        1.6       &       38.6     \\
        & HM    &       32      &  60.2   &  71.6   &   33.9  & 51.2    & \textbf{ 19.3}  &         25.9       &        19.6        &       25.7         &       0.65       &     0.45        &         30.3       &         36.7        &\textbf{46.0}&        8.8       &       46.1     \\
        & Ours  &  \textbf{16 } & {60.7}  &  71.2   &  34.6   & 49.4    &  17.7           &   \textbf{20.3}    &   \textbf{23.5}    &   \textbf{12.9}    &       0.96       &   \textbf{0.13} &   \textbf{39.1}    &   \textbf{40.6}     &    36.9     &    \textbf{13.7} &  \textbf{47.7} \\
  \cmidrule(lr){1-18}
    \rowcolor{gray!20}  \cellcolor{white} \multirow{4}*{SOP} 
        & N/A   &     N/A       &  {62.9}  & {68.5}  &  {39.2} & {87.4} &       0.1       &        99.3        &       0.2          &        99.1        &        0.85      &       1.69      &         0.0        &     0.0       &     6.3         &        0.0       &      {4.0}    \\
        & ACT   &     32        &  47.5    &  52.6   &   25.5  & 84.9   &      24.1       &        10.5        &       22.7         &        9.4         &  \textbf{0.25}   &      0.53       &         21.2       &     21.6      &     27.8        &        15.3      &      50.8     \\
        & HM    &     32        &  46.8    &  51.7   &   24.5  & 84.7   &       32.0        &        4.2         &       33.7         &        3.0         &         0.61     &      0.20       &         39.1       &     39.8      &     37.9        &        45.6      &      61.6     \\
        & Ours  & \textbf{16}   &  {48.3}  &  53.3   &  25.9   & 84.9   &  \textbf{32.3}  &    \textbf{3.7}    &   \textbf{36.0}    &   \textbf{2.6}     &        0.80     &   \textbf{0.14} &   \textbf{43.2}    & \textbf{45.1} &  \textbf{39.8}  &   \textbf{53.1}  &  \textbf{62.4}\\
    \bottomrule
    \end{tabular}
    }
    
\end{table*}

However, there are issues in ERS that make it infeasible for reasonable robustness evaluation:
(1) \textit{Inconsistent similarity metrics.} ERS includes TMA \citep{TMA}, the only cosine-similarity-based attack, as one of the attacks for robustness evaluation, which is not appropriate for evaluating models trained in the Euclidean space. Experimental results also indicate that TMA yields contradictory results against all other attacks (see Table.\ref{tabl_ERS} and Table.6 in \citep{HM}). (2) \textit{Initial state variations.} ERS is calculated using after-attack results, without considering the variation in the initial states before attacks. For example, in the CA+ attack (which raises the rank of a candidate over a query), despite the \textbf{randomly} selected targets, ERS ignores the variation in their before-attack ranks, \textit{e.g.}, $50\pm1$. This introduces noise into ERS and makes it inaccurate.
Similar issues persist for Recall attacks: since ERS calculates a joint total score, clean R@1 cannot become a reference for justification. 

\noindent \textbf{Adversarial Resistance Score.}
To solve the issues above, we remove TMA from evaluation for fairness consideration, and ES:D is also removed as it only considers attack-incurred embedding shifts and does not necessarily equal robustness. Specifically, the Adversarial Resistance Score (ARS) of an attack $\mathcal{A}$ against a model $\mathbf{M}$ is calculated based on the actual impact it makes compared to the intention of the attack, rather than directly using the results, defined as follows:
\begin{equation}
    \label{ar}
    \mathbb{R}_{\mathbf{M},\mathcal{A}} = (1 - \frac{ \mathcal{O}_{r} - \mathcal{O}_{i}}{ \mathcal{O}_{g} - \mathcal{O}_{i}}) \times 100\%,
\end{equation}
where  $\mathcal{O}_{i}$ refers to the initial value before attacks, $\mathcal{O}_{r}$ is the actual attacking result, and $\mathcal{O}_{g}$ is the intended result of the attack, respectively. 

Detailed calculations of ASR for different types of attacks can be found in Appendix \ref{ARS}.

Unlike ERS, our ARS calculates the percentage of change over the attack's intention, eliminating the variation in initial conditions for a more reasonable robustness evaluation.

\section{Experiments} 
In this section, we conduct comprehensive experiments to demonstrate the effectiveness of our CA-TRIDE, involving its comparison to other defense baselines, the validation of CA in preventing model collapse and TRIDE in solving the weak adversary, and ablation studies on the main components of our CA-TRIDE.

\subsection{Experimental Settings} \label{settings}
\noindent\textbf{Models and datasets.}
We follow the setting of \citet{ACT} and use a pre-trained ResNet-18 \citep{Res18} with the last layer changed to \textit{N}=512 as our baseline model. The triplet margin $\beta_{\mathcal{T}}$ is set as 0.2 for all datasets. Evaluations are on three popular datasets in image retrieval tasks, i.e. CUB-200-2011 \citep{CUB}, Cars-196 \citep{CARS}, and SOP \citep{lifted:ref}. We train our models using ADAM\citep{ADAM} optimizer with a $1.0\times 10^{-3}$ learning rate, a mini-batch size of 112, and training epochs of 100 under the above three datasets. For the top-rank pair, $\gamma = 0.5$ and the triplet margin in $L_{TR}$ $\beta_{TR}$ is 0.04.

\noindent\textbf{Adversaries.} Adversarial perturbation is generated through PGD \citep{PGD} with an optimization step $\alpha = 1/255$, 16 iterations and clipped by an $l_{\infty}$ norm of $\epsilon = 8/255$. A progressive PGD step size $\alpha$ is also deployed to help the model balance between accuracy and robustness. All CA-TRIDE implementation details are the same unless specified. Details are given in Section \ref{imp} of the Appendix. 

\noindent\textbf{Metrics.} Benign results are given as R@1, R@2, mAP, and NMI following the setting in \citet{HM}, while robustness scores are calculated using both the conventional metric, Empirical Robustness Score (ERS)\textcolor{black}{ \cite{ACT}}, and our proposed metric, Adversarial Resistance Score (ARS).

\subsection{CA-TRIDE vs. Other Defense Methods} \label{sec_jus}
We compare our CA-TRIDE to the existing SOTA methods, HM \citep{HM} and ACT \citep{ACT}, regarding their performance on both benign and adversarial examples, based on both ARS and ERS.
Results for ARS are presented in Table \ref{tabl_AR}, and ERS results are given in Table \ref{tabl_ERS}.
From both tables, we can draw the following conclusions: (1) CA-TRIDE significantly outperforms HM and ACT in almost all attacks under both metrics, which demonstrates the effectiveness of our CA-TRIDE. This suggests that \textbf{CA-TRIDE models can undertake stronger AT without model collapse}. (2) CA-TRIDE uses only half the PGD steps of the previous methods but exhibits higher robustness, at the cost of little or no drop in benign performances. This implies that \textbf{our TRIDE settings provide stronger adversaries and a more efficient AT paradigm.}

\begin{figure}[!t]
    \includegraphics[width=\columnwidth]{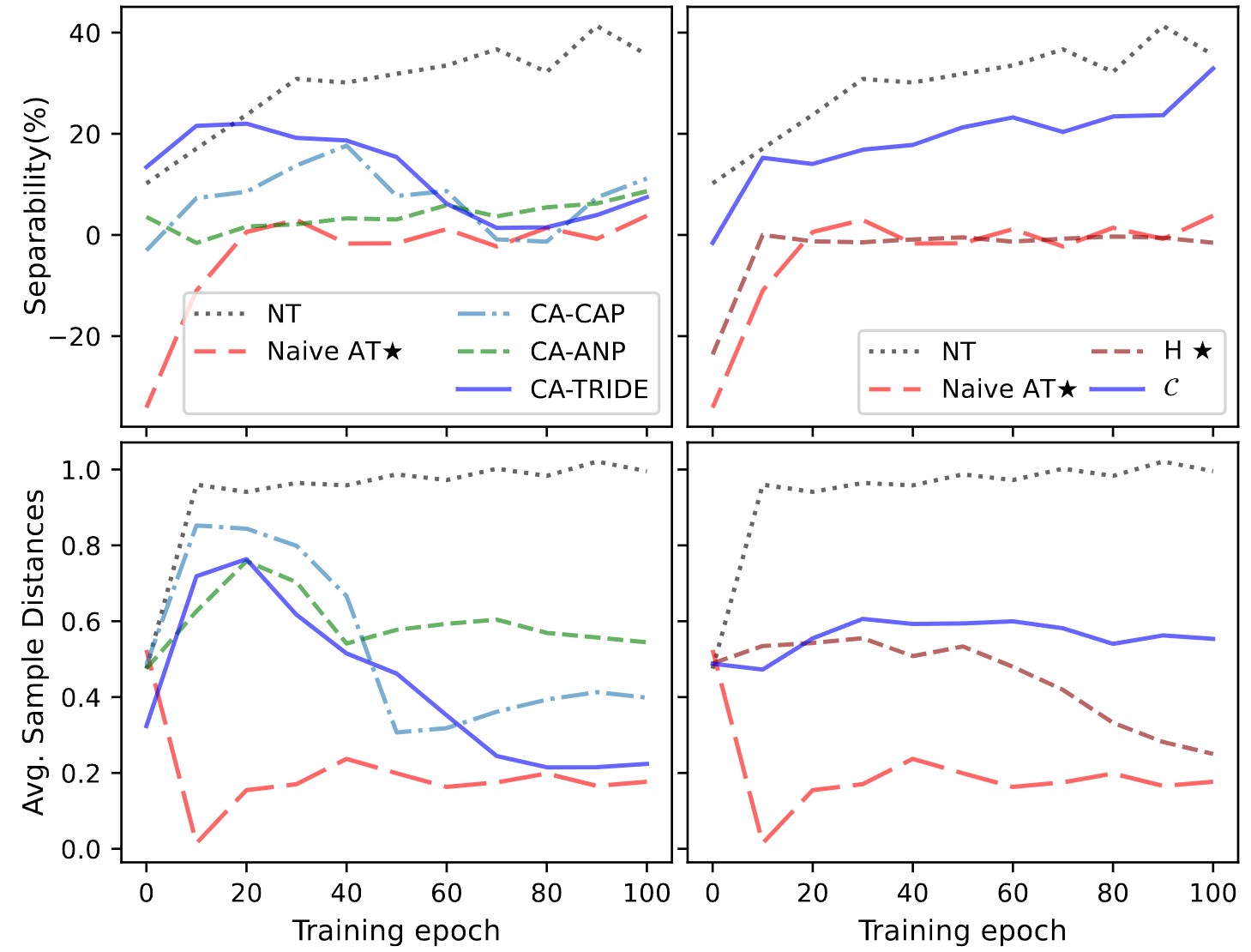}
    \vspace{-10pt}
\caption{\label{img_embchange}\textbf{Left Column:} Our TRIDE leads to lower separability and larger shrunk embedding distances compared with CAP/ANP, but the model remains uncollapsed due to our CA. \textbf{Right Column:} Our collapseness $\mathcal{C}$ outperforms hardness $H$ in preventing model collapse. $\bigstar$ denotes model collapse, i.e. separability $\approx0$ and heavily shrunk embeddings. The dataset is CUB.
}
\vspace{-10pt}
\end{figure}

As for the relatively lower performance of our models in TMA scores, it does not necessarily mean lower performance because all models are trained in the Euclidean space while TMA is evaluated using cosine similarity. Moreover, it is noticeable that models with higher overall robustness (i.e. HM and ours) constantly score lower in TMA than less robust models (ACT), which contradicts all other attacks.
As for GTM, this reflects the trade-off between emphasizing local robustness (top-rank samples) and global robustness (overall samples). In CA-TRIDE, we propose a top-rank pair to emphasize local robustness. However, GTM naively pushes the anchor towards the \textit{top-1 closest} negative example rather than a subset of top-rank samples, which contradicts the goal of our top-rank pair that prioritizes top-half negatives, thus leading to relatively lower scores. {The weakness of GTM is also implied by the non-zero attacking results on undefended models in Table \ref{tabl_AR} and Table \ref{tabl_ERS}.}

\subsection{Validation of CA and TRIDE}\label{validation}
In this section, we provide analyses to validate the individual effectiveness of our CA and TRIDE.
To this end, we train the following models: an NT model trained on vanilla data, a Naive AT model trained using a conventional adversary, a CA-CAP model, a CA-ANP model, and a CA-TRIDE model (i.e. our full implementation).

\begin{figure}[!t]
    \centering
\includegraphics[width=0.7\columnwidth]{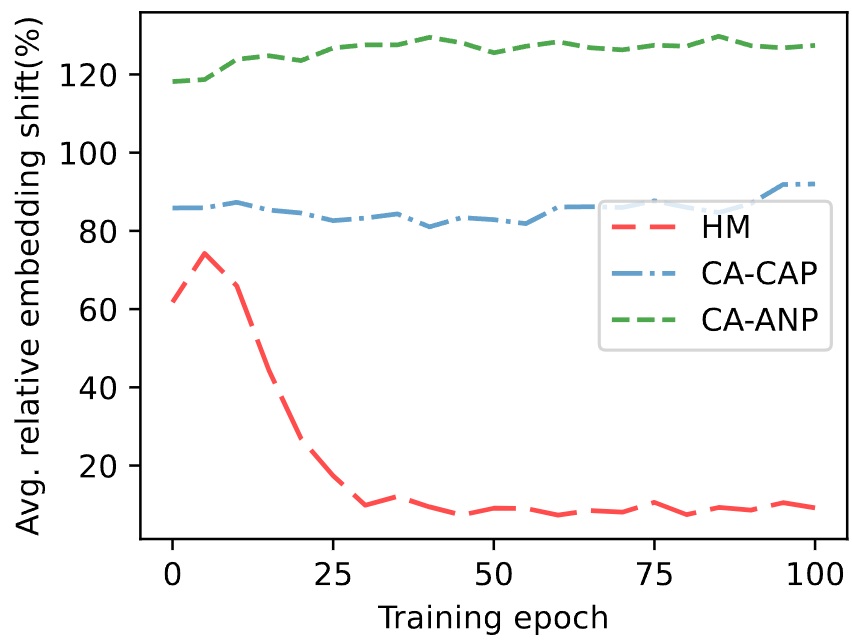}
    \vspace{-10pt}
    \caption{\label{img_embcomp} Our CA-ANP and CA-CAP cause substantially larger embedding shifts than HM. The dataset is CUB.}
    \vspace{-5pt}
\end{figure}

\noindent\textbf{CA prevents model collapse.}
To visualize model states, we calculate the per mini-batch average sample distances $\Bar{d}$ and use \textit{separability}, defined as $\frac{d(a,n)-d(a,p)}{\Bar{d}}$, to evaluate how well the model separates positive and negative samples. In particular, normalization by $\Bar{d}$ eliminates the influence of overall embedding changes. 

As shown in the left column of Figure \ref{img_embchange}, benign training (black dotted line) maximizes the separation between positives and negatives, without incurring a decrease in average sample distances (i.e. $\Bar{d}$). However, naive AT yields a drastic decrease in $\Bar{d}$ and harms separability by making it largely negative and fluctuating around 0. We thus summarise \textit{embedding shrinkage} and \textit{entangled samples} as two representative manifestations of model collapse.
The separability of CA-CAP and CA-ANP remains positive during training, and $\Bar{d}$ undergoes a mild shrinkage. This shows the effectiveness of collapse awareness in stopping model collapse, i.e. preventing embedding shrinkage and entangled samples.

\begin{table*}[!thbp]
    \caption{{Ablation study on CA-TRIDE vs. CA-CAP/-ANP. The dataset is CUB.}}
    \footnotesize
    \small
    \centering
   \resizebox{\linewidth}{!}{
    \begin{tabular}{cc|c|cccc|cccc|c|c}
    \toprule
    \multicolumn{2}{c|}{\textbf{Defense}} & \multirow{2}*{R@1$\uparrow$}  & \multicolumn{8}{c|}{Adversarial Example Evaluation (ARS scores)} &  {Overall} &   {Overall}   \\
    \cmidrule(lr){4-11}
    \multicolumn{2}{c|}{\textbf{Method}}& & CA+$\uparrow$ & CA-$\uparrow$ & QA+$\uparrow$ & QA-$\uparrow$ & ES:R $\uparrow$& LTM$\uparrow$ & GTM $\uparrow$& GTT $\uparrow$ &{ERS}$\uparrow$ & {ARS} (\%) $\uparrow$ \\
   
    \cmidrule(lr){1-13}
            \multicolumn{2}{c|}{CA-ANP}& 34.2  & 27.4 & 56.7 & 35.6 & {73.3} & {57.3} & {61.1} & {65.8} & {5.1} &  34.0 & 47.8\\
            \multicolumn{2}{c|}{CA-CAP}& 33.8 & {34.2} & {68.0} & {52.2} & {70.8} & 51.2 & 47.6 & 60.7 & 3.1 & {37.9} & {48.5} \\
            \multicolumn{2}{c|}{ CA-TRIDE} &   {34.9}   & {32.6}    & {68.5}     &   {41.8}    &   {79.2}    & {61.9}      &  {59.0}    &   64.8 &   {5.1} & \textbf{38.6} &\textbf{51.6} \\  
   \bottomrule
    \end{tabular}
    }
     
  \label{Tabl_ablation}
\end{table*}

\begin{table*}[!thbp]
 \caption{{Ablation study on the top-rank pair (TPR), i.e. $\Delta_{TR}$+$L_{TR}$. The dataset is CUB.}}
    \footnotesize
    \small
    \centering
   \resizebox{\linewidth}{!}{
    \begin{tabular}{cc|c|cccc|cccc|c|c}
    \toprule
    \multicolumn{2}{c|}{ {Defense}} & \multirow{2}*{R@1$\uparrow$}  & \multicolumn{8}{c|}{Adversarial Example Evaluation (ARS scores)} &   {Overall} &   {Overall}   \\
    \cmidrule(lr){4-11}
    \multicolumn{2}{c|}{{Method}}& & CA+$\uparrow$ & CA-$\uparrow$ & QA+$\uparrow$ & QA-$\uparrow$ & ES:R $\uparrow$& LTM$\uparrow$ & GTM $\uparrow$& GTT $\uparrow$ &{ERS}$\uparrow$ & {ARS} (\%) $\uparrow$ \\
   
    \cmidrule(lr){1-13}
           \cmidrule(lr){1-13}
           \multicolumn{2}{c|}{ w/o TPR} & 32.7  & 27.8 & 60.2 & 36.6 & 69.1 & 46.2 & 26.3 & 47.4 & 0.7 &  33.3 & 39.3 \\     
            \multicolumn{2}{c|}{w/o $L_{TR}$} & {34.6}  & \textbf{34.6}  &  \textbf{70.0}   &  \textbf{43.4}   &   {79.1}  &  57.8  & 54.9 &   {60.7} &  4.0 & 38.4 & 50.6 \\

        \cmidrule(lr){1-13}
            \multicolumn{2}{c|}{CA-TRIDE} &   {34.9}   & {32.6}    &  {68.5}     &   {41.8}    &   \textbf{79.2}    & \textbf{61.9}      &  \textbf{59.0}    &   \textbf{64.8} &   \textbf{5.1} & \textbf{38.6} &\textbf{51.6} \\  
   \bottomrule
    \end{tabular}
    }
    
  \label{Tabl_tr_ablation}
\end{table*}

Finally, to validate the superiority of $\mathcal{C}$ as a novel metric to track model states proactively, we train two models using CA-TRIDE but with different metrics: our collapse-aware adversary and hardness-aware adversary, denoted as $\mathcal{C}$ and H, respectively. Progressive $\alpha$ is disabled for a fairer comparison. Results are given in the right column of Figure \ref{img_embchange}. The behavior of the H-oriented model resembles that of Naive AT, both of which confront model collapse. The C-oriented model, on the other hand, behaves similarly to NT, implying the superiority of our $\mathcal{C}$.

\noindent\textbf{TRIDE solves the weak adversary.} 
To validate the effectiveness of our TRIDE, we compare the averaged embedding shifts caused by perturbations generated using three methods: HM, CAP, and ANP. All values are normalized similarly as separability to eliminate the influence of embedding shrinkage. As shown in Figure \ref{img_embcomp}, our CA-CAP and CA-ANP consistently yield larger embedding shifts than HM, especially in the later stage of the training. Our TRIDE surpasses HM regarding both average and overall embedding shifts. Here the overall embedding shift equals the averaged embedding shift multiplied by its corresponding number of perturbed components, i.e., 1 for ANP, 2 for CAP, and 3 for HM. This demonstrates the efficacy of TRIDE in maximizing embedding shifts. Qualitative analysis also aligns with this result, as discussed in Appendix \ref{IME}.
To further investigate the individual effectiveness of TRIDE, we train three model variants without using CA, i.e. a naive CAP model, a naive ANP model, and a naive TRIDE model.
Their R@1 results are 4.4\%, 7.8\% and 0.8\% respectively, indicating that, as expected, both CAP/ANP/TRIDE could yield strong adversaries sufficient to cause model collapse.

\begin{figure}[!t]
    \centering
    \includegraphics[width=\columnwidth]{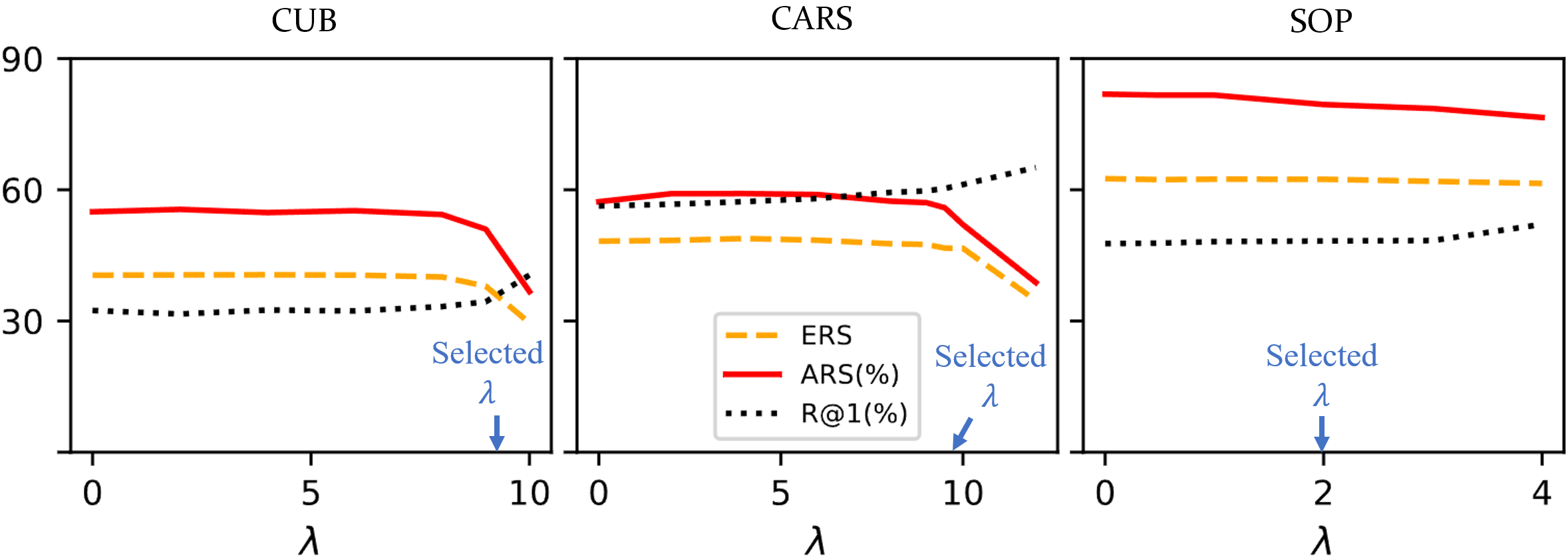}
    \vspace{-10pt}
\caption{\label{lambda_change}Ablation study on the attention factor $\lambda$.}
\vspace{-5pt}
\end{figure}

\subsection{Ablation Studies}\label{ablation}
In this section, we verify the effectiveness of CA-TRIDE by providing the ablation experiment results in Table \ref{Tabl_ablation} and \ref{Tabl_tr_ablation}.

\noindent\textbf{TRIDE vs CAP/ANP.} As shown in Table \ref{Tabl_ablation}, TRIDE does not necessarily reduce the overall strength of AT, and CA-CAP/ANP models emphasize differently on robustness. CA-CAP exhibits better robustness in almost all ranking attacks than CA-ANP, while the latter is more robust against Recall attacks, aligning with our intention of designing CAP and ANP. Finally, combining CAP and ANP yields well-rounded robustness across all attacks and achieves the best overall robustness.

\begin{figure}[!t]
    \centering
\includegraphics[width=0.7\columnwidth]{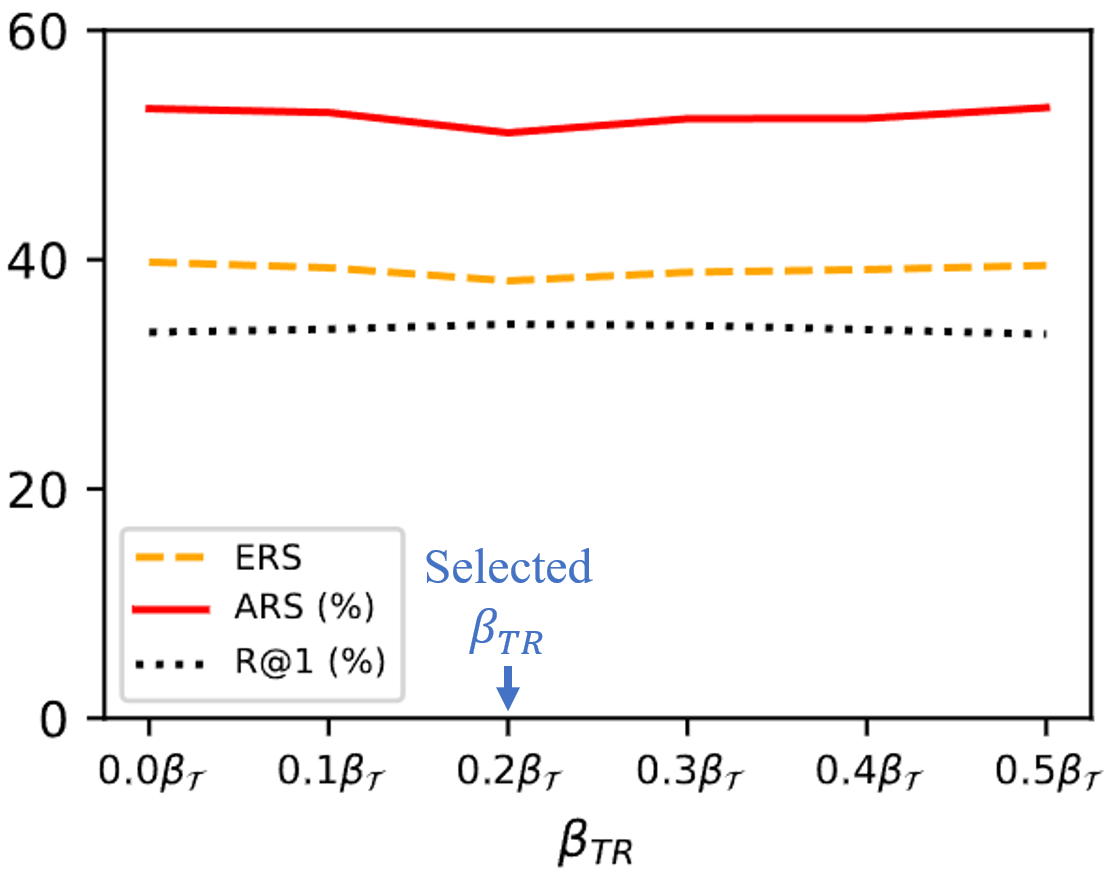}
    \vspace{-10pt}
    \caption{\label{fig_l_beta}Ablation study on the triplet loss margin $\beta_{TR}$ in $L_{TR}$. The dataset is CUB.}
    \vspace{-5pt}
\end{figure}

\noindent\textbf{Top-rank pair.}
As discussed in Section \ref{PSe}, the top-rank pair is incorporated to reinforce local corruption in ANP and help the model capture top-rank robustness against Recall attacks. According to the results presented in Table \ref{Tabl_tr_ablation}, the model with only $\Delta_{TR}$ already gains over 10\% of robustness boost against all attacks compared to its counterpart without our top-rank pair, with an impressive enhancement of performance on clean examples. This indicates the novelty of $\Delta_{TR}$ regarding the AT efficiency. Furthermore, when paired with the top-rank triplet $L_{TR}$, our model further acquires a noticeable increase in robustness against all Recal attacks, but at the cost of a marginal drop w.r.t. ranking attacks, implying a trade off between global and local robustness.

\noindent\textbf{Hyperparameters.} 
\noindent We also investigate how our models behave under different hyperparameters, i.e. the attention factors $\lambda$ in $\mathcal{C}$ and the triplet loss margin $\beta_{TR}$ in $L_{TR}$. $\lambda$ determines how much attention $\mathcal{C}$ should pay to anchor-proximity samples, as introduced in Equation \ref{eq_c}. Larger $\lambda$ means more focus on samples closer to anchors and less focus on the other samples. Note that when $\lambda=0$, our $\mathcal{C}$ falls back to $H$ by treating all examples identically. Results are presented in Figure \ref{lambda_change}. Overall, our CA-TRIDE is insensitive to $\lambda$, while a too-large $\lambda$ could further increase the clean accuracy but harm the robustness.
Hence, we choose the value of $\lambda$ that yields the closest R@1 performance to HM for a fair comparison, i.e. 10 for CUB, 9.5 for CARS, and 2.0 for SOP. 

{Moreover, as shown in Table \ref{Tabl_stas}, we find our settings of $\lambda$ on different datasets are correlated with their levels of entanglement, i.e., $d_{intra}/d_{inter}$, where $d_{intra}$ ($d_{inter}$) is the average intra (inter)-class distance of CA-TRIDE trained models. This correlation aligns with the intuition that more entangled datasets (i.e., CUB and CARS) require more attention on the anchor-proximity samples to avoid model collapse, while less entangled datasets such as SOP do not require much attention to handle such samples.}

\begin{table}[t!]
 \caption{{ Statistical results of inter-/intra-class distances on all datasets, evaluated using our CA-TRDIE models. Inter/Intra quantifies the level of entanglement of data distribution.}}
    \centering
   \resizebox{\linewidth}{!}{
    \begin{tabular}{c|c|c|c|c}
    \toprule
     \multirow{2}*{{Dataset}}& {Inter-Class}  & {Intra-Class}  &  \multirow{2}*{Entanglement}& \multirow{2}*{$\lambda$} \\
                             & {Distances}  & {Distances}  & &   \\
       \cmidrule(lr){1-5}
       CUB & 0.287 & 0.226 & 0.79 & 10.0 \\     
       CARS & 0.325 & 0.256 & 0.79 & 9.5 \\     
       SOP & 0.664 & 0.438 & 0.66 & 2.0 \\      
   \bottomrule
    \end{tabular}
    }
     \vspace{-5pt}
  \label{Tabl_stas}
\end{table}

For $\beta_{TR}$ in $L_{TR}$, it functions similarly to the $\beta_{\mathcal{T}}$ in the triplet loss: determining how hard the model should keep top-rank positives and negatives separated. Likewise, $L_{TR}$ is essentially a triplet loss variant calculated using only top-rank samples instead of all samples, and we thus evaluate the $\beta_{TR}$ as a proportion of $\beta_{\mathcal{T}}$, ranging from 0 to 50\%. As shown in Figure \ref{fig_l_beta}, no significant changes in robustness are found as $\beta_{TR}$ varies. Therefore, our method is also insensitive to the hyperparameter $\beta_{TR}$. We thus choose the value of $\beta_{TR}$ that yields the closest R@1 performance as HM for a fair comparison, i.e. $\beta_{TR}$ = 0.2$\beta_{\mathcal{T}}$.

\subsection{Training Time and Computational Cost}\label{TCE}

Since CA-TRIDE introduces extra operations during AT, we further compare it to HM regarding training time and computational cost. Results imply that both are reduced. We conducted 5 runs of HM and CA-TRIDE on the CUB dataset with an RTX3090 GPU. On average, CA-TRIDE takes 15\% less training time: 470 vs 550 minutes. We attribute this efficiency increase of CA-TRIDE to its triplet decoupling and the halved PGD steps, i.e., 16 (CA-TRIDE) vs. 32 (HM). 
For the computational cost, since CAP and ANP only use 2/3 (P and N) and 1/3 (A) of the triplets respectively, our CA-TRIDE yields only a $\frac{1}{2}\times(\frac{1}{3}+\frac{2}{3})=\frac{1}{2}$ computational cost of HM (and ACT).

\subsection{Compatibility of CA-TRIDE}

Our CA-TRIDE is a plug-and-play method for other triplet-based deep metric learning, regardless of datasets, models, and methods. Specifically, our CA-TRIDE does not change adversarial training at the system level but at the data sample level, by simply applying a weight to all samples (via our CA) and decoupling the perturbation update on them (via our TRIDE). In particular, it is worth mentioning that the only tunning possibly required is to find the attention factor $\lambda$ for the new dataset or model (if applicable) because as validated in Table \ref{Tabl_stas}, $\lambda$ is dataset-/model-dependent as it correlates with the level of entanglement.

\section{Limitations and Future Work}

\textbf{Globality vs locality.} {As presented in Table \ref{tabl_AR} and \ref{tabl_ERS}, our CA-TRIDE achieves lower results than HM w.r.t. GTM attacks on both CUB and CARS. This might be due to the conflict between the design of our top-rank pair and the attacking mechanism of GTM. A more sophisticated design of the top-rank pair may potentially mitigate these gaps. Besides, to trade off globality and locality, we empirically choose the top-half positives and negatives as top-rank samples when designing $L_{TR}$. It is possible to explore further refinement to such a trade-off in the future.}

\noindent\textbf{Attention factor $\lambda$.}
{Through our observation in Table \ref{Tabl_stas}, the level of entanglement, i.e., $d_{intra}/d_{inter}$, and our chosen attention factor $\lambda$ are found to be correlated. Therefore, a more rigorous and adaptive collapse-aware mechanism can be designed to adjust $\lambda$ better accordingly.}

\section{Conclusion}
\noindent In this paper, we have proposed CA-TRIDE, a novel approach to training image retrieval models with stronger adversarial robustness.
CA-TRIDE addresses two overlooked limitations: weak adversary and model collapse.
Specifically, TRIDE yields a strong adversary by spatially decoupling the optimization of perturbation on the triplets into ANP and CAP, while CA captures intermediate model states using a novel metric called \textit{collapseness} and integrates it into the subsequent optimization of perturbations. For evaluation metrics, we examine the conventional robustness evaluation metric ERS and identify two drawbacks. Consequently, we propose a new metric called ARS to address these drawbacks accordingly for reasonable robustness evaluations.
Extensive experiments on three datasets validate the superiority of our CA-TRIDE in both ERS and ARS. 

\section*{Acknowledgments}
This research is supported by the National Key Research and Development Program of China (2021YFB3100700), the National Natural Science Foundation of China (62376210, 62161160337, 62132011, U21B2018, U20A20177, 62206217), the Shaanxi Province Key Industry Innovation Program (2023-ZDLGY-38, 2021ZDLGY01-02).

\section*{Impact Statement}
This paper presents work that aims to advance the field of adversarial defense in image retrieval. Potential societal consequences of our work are less influential and have become scientific consensus, none of which we feel must be specifically highlighted here.

\bibliography{example_paper}

\begin{thebibliography}{34}
\providecommand{\natexlab}[1]{#1}
\providecommand{\url}[1]{\texttt{#1}}
\expandafter\ifx\csname urlstyle\endcsname\relax
  \providecommand{\doi}[1]{doi: #1}\else
  \providecommand{\doi}{doi: \begingroup \urlstyle{rm}\Url}\fi

\bibitem[Bai et~al.(2020)Bai, Chen, Li, Wu, Guo, Xia, and Yang]{AA:hash1}
Bai, J., Chen, B., Li, Y., Wu, D., Guo, W., Xia, S., and Yang, E.
\newblock Targeted attack for deep hashing based retrieval.
\newblock \emph{CoRR}, abs/2004.07955, 2020.
\newblock URL \url{https://arxiv.org/abs/2004.07955}.

\bibitem[Chen et~al.(2021)Chen, Lu, Wang, Qin, and Wang]{DAIR}
Chen, M., Lu, J., Wang, Y., Qin, J., and Wang, W.
\newblock Dair: A query-efficient decision-based attack on image retrieval systems.
\newblock In \emph{Proceedings of the 44th International ACM SIGIR Conference on Research and Development in Information Retrieval}, SIGIR '21, pp.\  1064–1073, New York, NY, USA, 2021. Association for Computing Machinery.
\newblock ISBN 9781450380379.
\newblock \doi{10.1145/3404835.3462887}.
\newblock URL \url{https://doi.org/10.1145/3404835.3462887}.

\bibitem[Feng et~al.(2020)Feng, Chen, Dai, and Xia]{PQAG}
Feng, Y., Chen, B., Dai, T., and Xia, S.
\newblock Adversarial attack on deep product quantization network for image retrieval.
\newblock \emph{CoRR}, abs/2002.11374, 2020.
\newblock URL \url{https://arxiv.org/abs/2002.11374}.

\bibitem[He et~al.(2015)He, Zhang, Ren, and Sun]{Res18}
He, K., Zhang, X., Ren, S., and Sun, J.
\newblock Deep residual learning for image recognition.
\newblock \emph{CoRR}, abs/1512.03385, 2015.
\newblock URL \url{http://arxiv.org/abs/1512.03385}.

\bibitem[Kingma \& Ba(2014)Kingma and Ba]{ADAM}
Kingma, D.~P. and Ba, J.
\newblock Adam: A method for stochastic optimization.
\newblock \emph{arXiv preprint arXiv:1412.6980}, 2014.

\bibitem[Krause et~al.(2013)Krause, Stark, Deng, and Fei-Fei]{CARS}
Krause, J., Stark, M., Deng, J., and Fei-Fei, L.
\newblock 3d object representations for fine-grained categorization.
\newblock In \emph{Proceedings of the IEEE international conference on computer vision workshops}, pp.\  554--561, 2013.

\bibitem[Li et~al.(2019)Li, Ji, Liu, Hong, Gao, and Tian]{UAP}
Li, J., Ji, R., Liu, H., Hong, X., Gao, Y., and Tian, Q.
\newblock Universal perturbation attack against image retrieval.
\newblock In \emph{2019 IEEE/CVF International Conference on Computer Vision (ICCV)}, pp.\  4898--4907, 2019.
\newblock \doi{10.1109/ICCV.2019.00500}.

\bibitem[Li et~al.(2021)Li, Li, Chen, Ye, He, Wang, Su, and Xue]{QAIR}
Li, X., Li, J., Chen, Y., Ye, S., He, Y., Wang, S., Su, H., and Xue, H.
\newblock {QAIR:} practical query-efficient black-box attacks for image retrieval.
\newblock \emph{CoRR}, abs/2103.02927, 2021.
\newblock URL \url{https://arxiv.org/abs/2103.02927}.

\bibitem[Liu et~al.(2019)Liu, Zhao, and Larson]{QAM}
Liu, Z., Zhao, Z., and Larson, M.~A.
\newblock Who's afraid of adversarial queries? the impact of image modifications on content-based image retrieval.
\newblock \emph{CoRR}, abs/1901.10332, 2019.
\newblock URL \url{http://arxiv.org/abs/1901.10332}.

\bibitem[Lu et~al.(2021)Lu, Chen, Sun, Wang, Wang, and Yang]{AA:hash2}
Lu, J., Chen, M., Sun, Y., Wang, W., Wang, Y., and Yang, X.
\newblock A smart adversarial attack on deep hashing based image retrieval.
\newblock In \emph{Proceedings of the 2021 International Conference on Multimedia Retrieval}, ICMR '21, pp.\  227–235, New York, NY, USA, 2021. Association for Computing Machinery.
\newblock ISBN 9781450384636.
\newblock \doi{10.1145/3460426.3463640}.
\newblock URL \url{https://doi.org/10.1145/3460426.3463640}.

\bibitem[Madry et~al.(2017)Madry, Makelov, Schmidt, Tsipras, and Vladu]{PGD}
Madry, A., Makelov, A., Schmidt, L., Tsipras, D., and Vladu, A.
\newblock Towards deep learning models resistant to adversarial attacks.
\newblock \emph{arXiv preprint arXiv:1706.06083}, 2017.

\bibitem[Musgrave et~al.(2020)Musgrave, Belongie, and Lim]{dmlcheck}
Musgrave, K., Belongie, S., and Lim, S.-N.
\newblock A metric learning reality check.
\newblock In \emph{Computer Vision--ECCV 2020: 16th European Conference, Glasgow, UK, August 23--28, 2020, Proceedings, Part XXV 16}, pp.\  681--699. Springer, 2020.

\bibitem[Oh~Song et~al.(2016)Oh~Song, Xiang, Jegelka, and Savarese]{lifted:ref}
Oh~Song, H., Xiang, Y., Jegelka, S., and Savarese, S.
\newblock Deep metric learning via lifted structured feature embedding.
\newblock In \emph{Proceedings of the IEEE conference on computer vision and pattern recognition}, pp.\  4004--4012, 2016.

\bibitem[Pang et~al.(2020)Pang, Yang, Dong, Xu, Zhu, and Su]{AT:ref4}
Pang, T., Yang, X., Dong, Y., Xu, K., Zhu, J., and Su, H.
\newblock Boosting adversarial training with hypersphere embedding.
\newblock \emph{Advances in Neural Information Processing Systems}, 33:\penalty0 7779--7792, 2020.

\bibitem[Picot et~al.(2022)Picot, Messina, Boudiaf, Labeau, Ayed, and Piantanida]{AT:ref3}
Picot, M., Messina, F., Boudiaf, M., Labeau, F., Ayed, I.~B., and Piantanida, P.
\newblock Adversarial robustness via fisher-rao regularization.
\newblock \emph{{IEEE} Transactions on Pattern Analysis and Machine Intelligence}, pp.\  1--1, 2022.
\newblock \doi{10.1109/tpami.2022.3174724}.
\newblock URL \url{https://doi.org/10.1109\%2Ftpami.2022.3174724}.

\bibitem[Roth et~al.(2020)Roth, Milbich, Sinha, Gupta, Ommer, and Cohen]{revisit}
Roth, K., Milbich, T., Sinha, S., Gupta, P., Ommer, B., and Cohen, J.~P.
\newblock Revisiting training strategies and generalization performance in deep metric learning.
\newblock In \emph{International Conference on Machine Learning}, pp.\  8242--8252. PMLR, 2020.

\bibitem[Schroff et~al.(2015)Schroff, Kalenichenko, and Philbin]{trip:ref}
Schroff, F., Kalenichenko, D., and Philbin, J.
\newblock Facenet: {A} unified embedding for face recognition and clustering.
\newblock \emph{CoRR}, abs/1503.03832, 2015.
\newblock URL \url{http://arxiv.org/abs/1503.03832}.

\bibitem[Sohn(2016)]{Npair:ref}
Sohn, K.
\newblock Improved deep metric learning with multi-class n-pair loss objective.
\newblock \emph{Advances in neural information processing systems}, 29, 2016.

\bibitem[Szegedy et~al.(2013)Szegedy, Zaremba, Sutskever, Bruna, Erhan, Goodfellow, and Fergus]{AA}
Szegedy, C., Zaremba, W., Sutskever, I., Bruna, J., Erhan, D., Goodfellow, I., and Fergus, R.
\newblock Intriguing properties of neural networks.
\newblock \emph{arXiv preprint arXiv:1312.6199}, 2013.

\bibitem[Tolias et~al.(2019)Tolias, Radenovic, and Chum]{TMA}
Tolias, G., Radenovic, F., and Chum, O.
\newblock Targeted mismatch adversarial attack: Query with a flower to retrieve the tower.
\newblock \emph{CoRR}, abs/1908.09163, 2019.
\newblock URL \url{http://arxiv.org/abs/1908.09163}.

\bibitem[Wang et~al.(2020)Wang, Wang, Li, Zhang, and Lin]{LTM}
Wang, H., Wang, G., Li, Y., Zhang, D., and Lin, L.
\newblock Transferable, controllable, and inconspicuous adversarial attacks on person re-identification with deep mis-ranking.
\newblock \emph{CoRR}, abs/2004.04199, 2020.
\newblock URL \url{https://arxiv.org/abs/2004.04199}.

\bibitem[Wang et~al.(2014)Wang, Song, Leung, Rosenberg, Wang, Philbin, Chen, and Wu]{IR}
Wang, J., Song, Y., Leung, T., Rosenberg, C., Wang, J., Philbin, J., Chen, B., and Wu, Y.
\newblock Learning fine-grained image similarity with deep ranking.
\newblock In \emph{Proceedings of the IEEE conference on computer vision and pattern recognition}, pp.\  1386--1393, 2014.

\bibitem[Wang et~al.(2017)Wang, Zhou, Wen, Liu, and Lin]{angular}
Wang, J., Zhou, F., Wen, S., Liu, X., and Lin, Y.
\newblock Deep metric learning with angular loss.
\newblock In \emph{Proceedings of the IEEE International Conference on Computer Vision (ICCV)}, Oct 2017.

\bibitem[Wang et~al.(2019)Wang, Han, Huang, Dong, and Scott]{Multsim:ref}
Wang, X., Han, X., Huang, W., Dong, D., and Scott, M.~R.
\newblock Multi-similarity loss with general pair weighting for deep metric learning.
\newblock In \emph{Proceedings of the IEEE/CVF conference on computer vision and pattern recognition}, pp.\  5022--5030, 2019.

\bibitem[Welinder et~al.(2010)Welinder, Branson, Mita, Wah, Schroff, Belongie, and Perona]{CUB}
Welinder, P., Branson, S., Mita, T., Wah, C., Schroff, F., Belongie, S., and Perona, P.
\newblock Caltech-ucsd birds 200.
\newblock Technical Report CNS-TR-201, Caltech, 2010.
\newblock URL \url{/se3/wp-content/uploads/2014/09/WelinderEtal10_CUB-200.pdf, http://www.vision.caltech.edu/visipedia/CUB-200.html}.

\bibitem[Wu et~al.(2017)Wu, Manmatha, Smola, and Krahenbuhl]{dw}
Wu, C.-Y., Manmatha, R., Smola, A.~J., and Krahenbuhl, P.
\newblock Sampling matters in deep embedding learning.
\newblock In \emph{Proceedings of the IEEE international conference on computer vision}, pp.\  2840--2848, 2017.

\bibitem[Xuan et~al.(2020)Xuan, Stylianou, Liu, and Pless]{MC}
Xuan, H., Stylianou, A., Liu, X., and Pless, R.
\newblock Hard negative examples are hard, but useful.
\newblock In Vedaldi, A., Bischof, H., Brox, T., and Frahm, J.-M. (eds.), \emph{Computer Vision -- ECCV 2020}, pp.\  126--142, Cham, 2020. Springer International Publishing.
\newblock ISBN 978-3-030-58568-6.

\bibitem[Yu et~al.(2018)Yu, Liu, Gong, Ding, and Tao]{trip:ref2}
Yu, B., Liu, T., Gong, M., Ding, C., and Tao, D.
\newblock Correcting the triplet selection bias for triplet loss.
\newblock In \emph{Proceedings of the European Conference on Computer Vision (ECCV)}, pp.\  71--87, 2018.

\bibitem[Zhang et~al.(2019)Zhang, Yu, Jiao, Xing, El~Ghaoui, and Jordan]{AT:ref2}
Zhang, H., Yu, Y., Jiao, J., Xing, E., El~Ghaoui, L., and Jordan, M.
\newblock Theoretically principled trade-off between robustness and accuracy.
\newblock In \emph{International conference on machine learning}, pp.\  7472--7482. PMLR, 2019.

\bibitem[Zhong \& Deng(2019)Zhong and Deng]{AT:ref1}
Zhong, Y. and Deng, W.
\newblock Adversarial learning with margin-based triplet embedding regularization.
\newblock In \emph{Proceedings of the IEEE/CVF international conference on computer vision}, pp.\  6549--6558, 2019.

\bibitem[Zhou \& Patel(2022)Zhou and Patel]{HM}
Zhou, M. and Patel, V.~M.
\newblock Enhancing adversarial robustness for deep metric learning.
\newblock In \emph{Proceedings of the IEEE/CVF Conference on Computer Vision and Pattern Recognition}, pp.\  15325--15334, 2022.

\bibitem[Zhou et~al.(2021{\natexlab{a}})Zhou, Wang, Niu, Zhang, Xu, Zheng, and Hua]{RA}
Zhou, M., Wang, L., Niu, Z., Zhang, Q., Xu, Y., Zheng, N., and Hua, G.
\newblock Practical relative order attack in deep ranking.
\newblock \emph{CoRR}, abs/2103.05248, 2021{\natexlab{a}}.
\newblock URL \url{https://arxiv.org/abs/2103.05248}.

\bibitem[Zhou et~al.(2021{\natexlab{b}})Zhou, Wang, Niu, Zhang, Zheng, and Hua]{ACT}
Zhou, M., Wang, L., Niu, Z., Zhang, Q., Zheng, N., and Hua, G.
\newblock Adversarial attack and defense in deep ranking.
\newblock \emph{CoRR}, abs/2106.03614, 2021{\natexlab{b}}.
\newblock URL \url{https://arxiv.org/abs/2106.03614}.

\bibitem[Zhu et~al.(2021)Zhu, Zhang, Han, Liu, Niu, Yang, Kankanhalli, and Sugiyama]{AT:ref5}
Zhu, J., Zhang, J., Han, B., Liu, T., Niu, G., Yang, H., Kankanhalli, M.~S., and Sugiyama, M.
\newblock Understanding the interaction of adversarial training with noisy labels.
\newblock \emph{CoRR}, abs/2102.03482, 2021.
\newblock URL \url{https://arxiv.org/abs/2102.03482}.

\end{thebibliography}
\bibliographystyle{icml2024}

\newpage
\appendix
\onecolumn
\section{Theoretical Analysis of Weak Adversary }\label{IME}
\noindent In this section, we provide a qualitative analysis of how and why the weak adversary leads to minimized embedding shifts, which necessitates our triplet decoupling (TRIDE) mechanism. 

As discussed in the main text of the paper, the current adversarial perturbation ${\delta}$ is acquired by:
\begin{equation}
\arg \max_{{\delta}} H(\Tilde{\mathbf{A}},\Tilde{\mathbf{P}},\Tilde{\mathbf{N}})
\end{equation}
Essentially, although the specific adversarial losses $L_{adv}$ vary across methods, the general goal of $L_{adv}$ is to maximize $H$ adversarially. Thus, we denote the consequential change caused by the perturbation as $\Delta_{H}$, defined as follows:
\begin{equation}
\label{l_adv}
    \Delta_{H} = H(\Tilde{\mathbf{A}},\Tilde{\mathbf{P}},\Tilde{\mathbf{N}}) - H({\mathbf{A}},{\mathbf{P}},{\mathbf{N}})
\end{equation}

As mentioned in the main paper, unlike AT in image classification, AT in DML has multiple choices for perturbation targets (anchors \textbf{A}, positives \textbf{P}, and negatives \textbf{N}). The average embedding shift of the perturbed targets depends on two factors: the angle between $\Bar{an}$ and $\Bar{ap}$, denoted as $\theta$ \cite{angular}, and the perturbation methods, denoted as $\mathcal{P}$. We will then determine how these factors influence the overall embedding shifts.

\begin{figure}[h]
    \centering
    \begin{minipage}{\linewidth} 
        \centering
        \subfigure[1][$\mathcal{P}=$ ANP, $\theta=\pi$]{\label{ANP_180}
            \includegraphics[width=0.18\linewidth]{180_ANP_v1.png}}
        \subfigure[2][$\mathcal{P}=$ CAP, $\theta=\pi$]{\label{CAP_180}
            \includegraphics[width=0.2\linewidth]{180_CAP_v1.png}}
        \subfigure[3][$\mathcal{P}=$ SIP, $\theta=\pi$]{\label{SIP_180}
            \includegraphics[width=0.2\linewidth]{180_SIP_v1.png}}
         \\
        \subfigure[b][$\mathcal{P}=$ ANP, $0<\theta<\pi$]{\label{ANP_o}
            \includegraphics[width=0.19\linewidth]{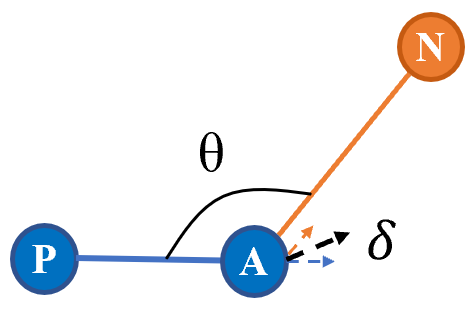}}
        \subfigure[d][$\mathcal{P}=$ CAP, $0<\theta<\pi$]{\label{CAP_o}
            \includegraphics[width=0.23\linewidth]{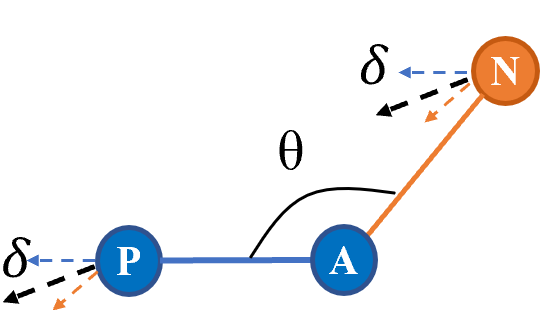}}
        \subfigure[f][$\mathcal{P}=$ SIP, $0<\theta<\pi$]{\label{SIP_o}
            \includegraphics[width=0.2\linewidth]{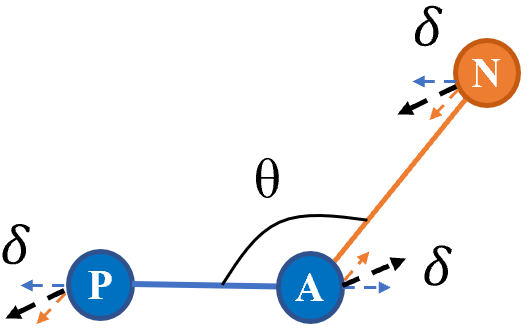}}
    \end{minipage}
    
    \caption{\label{theta} Demonstration on scenarios of different angle $\theta$ and perturbation method. $\Delta$ denotes the average embedding shifts of perturbed samples. ANP for anchor perturbation (only perturbs anchors), CAP for candidates perturbation (only perturbs candidates), and SIP for simultaneous perturbation (perturbs all the triplets). $\pi<\theta<2\pi$ is omitted due to symmetry.}
\end{figure}
\noindent As shown in Figure \ref{theta}, 
$\theta$ influences \textbf{how the perturbation changes the overall hardness}. In other words, with the same embedding shifts, $\theta$ determines the proportion of embedding shifts that transfer to $\Delta_{H}$. To calculate this proportion, we calculate a function $\gamma_{\theta}$ of $\theta$. Hence, given $\gamma(\theta)$ and an embedding shift $\delta$, $\Delta_{H}$ is given by:
\begin{equation}
    \Delta_H = \gamma_{\theta}\cdot \delta
\end{equation}
According to Figure \ref{theta}, $\gamma_{\theta}$ could range from 0 ($\theta = 0$, \textit{i.e.}, $\Delta_{H}= 0$) to 2 ($\theta = \pi$, \textit{i.e.}, $\Delta_{H} = 2\delta$.)

\subsection{Perturbation targets}

\noindent Perturbation methods $\mathcal{P}$ determine the total number of selected targets, (\textit{i.e.}, 1, 2 or 3), influencing {how the desired embedding shift $\Delta_{H}$ is allocated to all perturbation targets}. A general way to understand the idea is like assigning a certain amount of tasks (target hardness). The more people (samples) get assigned, the fewer tasks (embedding shifts) each person would have.

For analysis, we assume that all perturbed targets move identically by $\Delta$. In other words, $d_{a,\Tilde{a}} = \delta_{ANP}$ for anchor perturbation (ANP), and $d_{p,\Tilde{p}} = d_{n,\Tilde{n}} = \delta_{CAP}$ for candidate perturbation (CAP), and $d_{a,\Tilde{a}} = d_{p,\Tilde{p}} = d_{n,\Tilde{n}} = \delta_{SIP}$ for simultaneous perturbation (SIP), namely the existing method. 

\subsection{Calculations}

\noindent To calculate the averaged embedding shifts given $\theta$ and $\mathcal{P}$, we need to determine how much of each perturbed sample needs to move to achieve the required hardness increase $\Delta_{H} - H(\mathbb{T})$. We further drop the term $H(\mathbb{T})$ as it is identical for the same initial triplet. \textbf{Note that to simplify the calculation, we assume all perturbed samples have identical embedding shifts.}
We can now qualitatively demonstrate how different perturbation methods $\mathcal{P}$ and $\theta$ determine the averaged overall embedding shifts $\delta$ given the same $\Delta_{H}$, in a special-to-general manner:

\noindent (1) \textit{Special case}: $\theta = \pi$. For $\mathcal{P}=$ ANP, as shown in Figure \ref{ANP_180}, the gradient to increase $d(a,p)$ aligns with the gradient to decrease $d(a,n)$, which doubles the hardness shift caused by $\delta_{ANP}$. Thus, we can obtain the average embedding shift $\delta_{ANP}$ of ANP follows:
\begin{flalign}
    \gamma_{\theta}\cdot d_{a,\Tilde{a}} &=  \Delta_{H} \nonumber \\
    2\delta_{ANP} &=  \Delta_{H} \nonumber \\
    \delta_{ANP} &= \dfrac{\Delta_{H}}{2}
    \label{ANP_180_}
\end{flalign}
\noindent Turning to $\mathcal{P}=$ CAP, as demonstrated in Figure \ref{CAP_180}, $\gamma_{\theta} = 1$, and $\delta_{CAP} $ given as follows:
\begin{flalign}
    \gamma_{\theta}\cdot d_{p,\Tilde{p}} + \gamma_{\theta}\cdot d_{n,\Tilde{n}} =   \Delta_{H} \nonumber \\
    \delta_{CAP} + \delta_{CAP} =   \Delta_{H} \nonumber \\
    \delta_{CAP} =  \dfrac{ \Delta_{H}}{2}
    \label{CAP_180_}
\end{flalign}
For their combination, SIP leads to the least average embedding shift (as shown in Figure \ref{SIP_180}), which can be similarly calculated :
\begin{flalign}
    \gamma_{\theta}\cdot d_{a,\Tilde{a}}  + \gamma_{\theta}\cdot d_{p,\Tilde{p}} + \gamma_{\theta}\cdot d_{n,\Tilde{n}}  =  \Delta_{H} \nonumber \\
    2\delta_{SIP} +  \delta_{SIP} + \delta_{SIP} =  \Delta_{H} \nonumber \\
    \delta_{SIP} =  \dfrac{ \Delta_{H}}{4}
    \label{SIP_180_}
\end{flalign}
In this case, by comparing Equation \ref{ANP_180_}, Equation \ref{CAP_180_}, and Equation \ref{SIP_180_}, the overall average embedding shift of ANP and CAP is twice the average embedding shift of SIP.

\noindent (2) \textit{General cases:} $ 0< \theta < \pi$. For ANP, as shown in Figure \ref{ANP_o}, $\gamma_{\theta}$ becomes a function of $\theta$. Through geometric calculation, we obtain the approximation of $\gamma_{\theta}$, given as follows:
\begin{flalign}
    \gamma_{\theta} = 2cos(\dfrac{\pi - \theta}{2})
\end{flalign}
Then,  $\delta_{ANP}$ of ANP is calculated as:
\begin{flalign}
    \gamma_{\theta}\cdot d_{a,\Tilde{a}} = &\Delta_{H} \nonumber \\
    2cos(\dfrac{\pi - \theta}{2}) \cdot \delta_{ANP} = & \Delta_{H} \nonumber \\
    \delta_{ANP} =  & \dfrac{\Delta_{H}}{ 2cos(\dfrac{\pi - \theta}{2})}
    \label{ANP_o_}
\end{flalign}
This result can be verified by simply inserting $\theta = \pi$ into Equation \ref{ANP_o_}, which gives the same result in Equation \ref{ANP_180_}.

Similar to ANP, CAP can be regarded as applying an equivalent perturbation to \textbf{P} and \textbf{N}, with $\gamma_{\theta}$ becoming half of ANP, (\textit{i.e.}, half of $ 2cos(\dfrac{\pi - \theta}{2})$) :
\begin{flalign}
    \gamma_{\theta}\cdot d_{p,\Tilde{p}} + \gamma_{\theta}\cdot d_{n,\Tilde{n}} =  &\Delta_{H} \nonumber \\
   2\times cos(\dfrac{\pi - \theta}{2}) \cdot \delta_{CAP} =  &\Delta_{H} \nonumber \\
    \delta_{CAP} = & \dfrac{\Delta_{H}}{ 2cos(\dfrac{\pi - \theta}{2})}
    \label{CAP_o_}
\end{flalign}
For SIP, $\delta_{SIP}$ can once again be calculated by combining CAP and ANP:
\begin{flalign}
    \gamma_{\theta}\cdot d_{a,\Tilde{a}}  + \gamma_{\theta}\cdot d_{p,\Tilde{p}} + \gamma_{\theta}\cdot d_{n,\Tilde{n}}  =  &\Delta_{H} \nonumber \\
    4cos(\dfrac{\pi - \theta}{2})\delta_{SIP} =  &\Delta_{H} \nonumber \\
   \delta_{SIP} = &\dfrac{\Delta_{H}}{ 4cos(\dfrac{\pi - \theta}{2})}
    \label{SIP_o_}
\end{flalign}
Finally, we can compare SIP with CAP + ANP in more general cases ($0<\theta<\pi$) by comparing Equation \ref{ANP_o_}, Equation \ref{CAP_o_} and Equation \ref{SIP_o_}. Ideally, with the exactly identical $\theta$, $\Delta_{CAP,ANP}$ is still two times of $\delta_{SIP}$. However, in practical scenarios, there will be a phase difference most of the time. Hence, the magnitude ratio of CAP + ANP over SIP in more general cases can be given as follows:
\begin{flalign}
   \dfrac{\Delta_{CAP,ANP}}{\delta_{SIP}} = \dfrac{2cos({\dfrac{\pi - \theta_{1}}{2})}}{cos(\dfrac{\pi - \theta_{2}}{2})}
    \label{ratio}
\end{flalign}
For $0<\theta<\pi$, the denominator of the ratio can be regarded as a coefficient ranging from 0 to 1, with most of the time less than 1. This implies that this denominator mostly acts as an amplifier, making the overall ratio even larger. Due to symmetry, the scenario when $\pi<\theta<2\pi$ is identical to what has been discussed. In other words, despite some variation, Equation \ref{ratio} aligns with the experimental results shown in Figure \ref{img_embcomp}, which suggests that \textbf{our CAP + ANP setting outperforms the extant SIP regarding maximizing the average embedding shift under the same perturbation.}

\subsection{Conclusions}

Our analysis is also validated experimentally in Section 4.2. Our Tride consistently outperforms SIP w.r.t. the average embedding shift under perturbation, with the largest gap being almost 4 times the averaged embedding shifts of existing methods (SIP). 

Our theoretical analysis and experimental results demonstrate that the existing method leads to a significantly decreased perturbation-caused embedding shift compared to our TRIDE (CAP+ANP) setting. Consequently, the perturbation generated likewise is much weaker than the perturbation optimized using TRIDE. 

Another reason for TRIDE is that simultaneous perturbation is not a practical method considering many existing attacks against DML. In black-box scenarios, most existing attacks are achieved by ANP\cite{QAIR, DAIR, RA, UAP}, which is more practical and effective.

\section{Calculations Details of Adversarial Resistance Scores}
\label{ARS}

\noindent\textbf{Ranking attacks.} 
\noindent Ranking attacks intend to manipulate the rank of candidates through adversarial perturbations. Given a ranking attack $\mathcal{A}_{rank}$, for the $i$th candidate, we denote its initial rank $r$ for the $j$th query as $r_{i,j}$, and its after-attack rank as $\Tilde{r}_{i,j}$. Taking CA+ as an example, the goal of which is to elevate $r_{i,j}$ as much as possible, \textit{i.e.}, $r=0$. In other words, for this trial of CA+ attack, $\mathcal{O}_{g} = 0$, $\mathcal{O}_{i} = r_{ij}$ and $\mathcal{O}_{r} = \Tilde{r}_{i,j}$. Hence, $\mathbb{R}_{\mathbf{M},CA_+}$ can be calculated as follows:
\begin{flalign}
    \label{ar_ca}
    \mathbb{R}_{\mathbf{M},CA_+}& = (1 - \frac{ \mathcal{O}_{r} - \mathcal{O}_{i}}{ \mathcal{O}_{g} - \mathcal{O}_{i}}) \times 100\%, \nonumber \\
   & = \big( 1 - \frac{|r_{i,j} - \Tilde{r}_{i,j}|}{r_{i,j}} \big)\times 100\%  \\
\end{flalign} 
Similarly, AR all rank attacks (CA+, CA-, QA+, QA-) can be calculated likewise:
\begin{equation}
    \label{ar_rank}
    \mathbb{R}_{\mathbf{M},(CA_+, CA_-, QA_+, QA_-)} =  \big(1 - \frac{|r_{i,j} - \Tilde{r}_{i,j}|}{r_{i,j}}\big) \times 100\%  \\
\end{equation}

Note that the final ARS of a ranking attack is calculated by the average of $\mathbf{N}$ selected candidates (CA) or queries (QA), calculated as follows:
\begin{equation}
    \label{ar_rank_avg}
    \mathbb{R}_{\mathbf{M},\mathcal{A}_{rank}} =  \frac{1}{N}\sum^{\mathbf{N}}_{i,j}{\big(1 - \frac{|r_{i,j} - \Tilde{r}_{i,j}|}{r_{i,j}}\big) \times 100\%}  \\
\end{equation}
Details of CA and QA attacks can be found in \citep{ACT}.

\noindent\textbf{Recall attacks.}
\noindent The evaluation of Recall attacks is much simpler as these attacks seek to lower the R@1. Thus, given a model $\mathbf{M}$ with initial R@1 $\mu_{\mathbf{M}}$ and its after-attack R@1 $\Tilde{\mu}_{\mathbf{M}}$, all Rrecall attacks' intention is to lower $\mu_{\mathbf{M}}$ to 0, \textit{i.e.}, $\mathcal{O}_{g} = \mu_{\mathbf{M}}$, $\mathcal{O}_{a} = \Tilde{\mu}_{\mathbf{M}}$. Hence, $\mathbb{R}_{\mathbf{M},(ES,LTM,GTM)}$ can be calculated as follows:

\begin{flalign}
    \label{6}
   \mathbb{R}_{\mathbf{M},(ES,LTM,GTM)} &= \big(1 - \frac{\mu_{\mathbf{M}} - \Tilde{\mu}_{\mathbf{M}}}{\mu_{\mathbf{M}}}\big) \times 100\%  \nonumber \\
   & =\frac{\Tilde{\mu}_{\mathbf{M}}}{\mu_{\mathbf{M}}} \times 100\% 
\end{flalign}

In essence, our proposed ARS quantifies the robustness of a model based on how well the attack achieves its goal on this model. Initial state variation is eliminated by calculating the difference instead of evaluating pure results. 

\section{Implementation Details}
\label{imp}

\noindent To help the model balance between learning meaningful information and acquiring robustness, we apply a simple epoch-wise adjustment strategy to mini-batch sampling and perturbation adding.

For mini-batch sampling, we follow the semi-hard sampling \cite{trip:ref}, which samples hard examples within a given range $d(a,p) < d(a,n) < d(a,p) + \eta$, and apply an epoch-wise strategy to $\eta$: 
\begin{equation}
\label{gh}
\eta = \eta_0(1 - (\frac{n}{2\times n_{total}})^2)
\end{equation}
where $\eta_0$ is the preset margin of semi-hard sampling, $n$ stands for the current number of epochs, and $n_{total}$ stands for the total number of epochs.

For progressive step size, we similarly multiply the PGD step size $\alpha$ by $\frac{n}{n_{total}}$ to enable a gradually increasing adversary strength.


\end{document}